\documentclass[applsci,article,submit,pdftex,moreauthors]{mdpi}

\firstpage{1} 
\pubvolume{1}
\issuenum{1}
\articlenumber{0}
\pubyear{2025}
\copyrightyear{2025}
\datereceived{ } 
\daterevised{ } % Comment out if no revised date
\dateaccepted{ } 
\datepublished{ } 
\hreflink{https://doi.org/} % If needed use \linebreak

\usepackage{comment}
\usepackage{algorithm}
\usepackage{verbatimbox}
\usepackage{algorithmic}
\usepackage{amsfonts}
\usepackage{textcomp}
\usepackage{fancyvrb}
\usepackage{lettrine}
\usepackage{fancyhdr}
\usepackage{listings}
\usepackage{makecell}
\usepackage[most]{tcolorbox}
\usepackage[most]{tcolorbox}
\usepackage{listings}

\newtcblisting{rulebox}[2][]{
  colback=gray!3!white,     % softer background
  colframe=gray!50!black,  % lighter frame
  boxrule=0.4pt,           % thinner border
  arc=2pt,                 % small rounded corners
  top=1pt, bottom=1pt,     % tighter vertical padding
  left=2pt, right=2pt,     % tighter horizontal padding
  boxsep=1pt,              % less space inside box
  width=\linewidth,     % make box slightly smaller than text width
  listing only,
  title={#2},              % dynamic title
  fonttitle=\bfseries\footnotesize, % smaller title
  breakable,
  listing options={
    basicstyle=\ttfamily\footnotesize, % smaller font
    keywordstyle=\color{blue}\bfseries,
    commentstyle=\color{gray},
    morekeywords={:-,;,not},
    escapeinside={(*@}{@*)}
  },
  #1
}

\Title{Symbolic Imitation Learning: From Black-Box to Explainable Driving Policies}

\TitleCitation{Symbolic Imitation Learning: From Black-Box to Explainable Driving Policies}

\Author{Iman Sharifi $^{1}$\orcidA{}, Mustafa Yildirim $^{2,3,*}$\orcidB{} and Saber Fallah $^{2}$\orcidC{}}

\AuthorNames{Firstname Lastname, Firstname Lastname and Firstname Lastname}

\isAPAStyle{%
       \AuthorCitation{Sharifi, I., Yildirim, M., \& Fallah, S.}
         }{%
        \isChicagoStyle{%
        \AuthorCitation{Lastname, Firstname, Firstname Lastname, and Firstname Lastname.}
        }{
        \AuthorCitation{Sharifi, I., Yildirim, M., \& Fallah, S.}
        }
}

\address{%
$^{1}$ \quad Department of Science and Engineering, the George Washington University, Washington DC, USA; i.sharifi@email.gwu.edu\\
$^{2}$ \quad Connected and Autonomous Vehicles Lab, Department of Mechanical Engineering, University of Surrey, Guildford, UK; \{m.yildirim, s.fallah\}@surrey.ac.uk\\
$^{3}$ \quad Mechanical Engineering Department, Istanbul University-Cerrahpasa, Istanbul, Türkiye; mustafa.yildirim@iuc.edu.tr}
\abstract{Current imitation learning approaches, predominantly based on deep neural networks (DNNs), offer efficient mechanisms for learning driving policies from real-world datasets. However, they suffer from inherent limitations in interpretability and generalizability—issues of critical importance in safety-critical domains such as autonomous driving. In this paper, we introduce \textit{Symbolic Imitation Learning} (SIL), a novel framework that leverages Inductive Logic Programming (ILP) to derive explainable and generalizable driving policies from synthetic datasets. We evaluate SIL on real-world HighD and NGSim datasets, comparing its performance with state-of-the-art neural imitation learning methods using metrics such as collision rate, lane change efficiency, and average speed. The results indicate that SIL significantly enhances policy transparency while maintaining strong performance across varied driving conditions. These findings highlight the potential of integrating ILP into imitation learning to promote safer and more reliable autonomous systems.
}


\makeatletter
\let\linenumbers\relax
\makeatother

\begin{document}

\section{Introduction}

The development of autonomous driving technologies has elevated transportation systems to a new level, promising safer and more efficient roadways. Among the various techniques used to enable autonomous vehicles (AVs), imitation learning has emerged as a promising approach due to its ability to learn complex driving behaviors from expert demonstrations~\cite{codevilla2019exploring}. By leveraging large-scale driving datasets and deep neural networks (DNNs), imitation learning has demonstrated remarkable success in training autonomous agents to emulate human driving behaviors~\cite{pan2017agile, pan2020imitation, zhang2016query, morga2024behavioral, ho2016generative}.

Despite its impressive performance, DNN-based imitation learning (DNNIL), often implemented using Convolutional Neural Networks (CNNs) and Generative Adversarial Networks (GANs)~\cite{goodfellow2020generative}, inherits critical limitations from DNNs that hinder its widespread adoption in real-world autonomous systems. A prominent drawback is the lack of interpretability in the learned driving policies due to the black-box nature of DNNs~\cite{zheng2024explaining, zhang2021survey}, making the decision-making process of autonomous driving difficult to verify and validate. This lack of interpretability not only limits human ability to diagnose failures or errors but also hampers the trust and acceptance of autonomous systems by society and regulatory bodies~\cite{kim2017interpretable}.

Moreover, the generalizability of DNNIL driving policies remains a concern. Although these models can learn to imitate expert drivers in specific scenarios, adapting the learned policies to unseen situations can be problematic, as DNNIL is limited to behaviors observed during training~\cite{ghasemipour2020divergence}. Consequently, when there is a mismatch between test and training data distributions~\cite{zhu2020off}, the models often exhibit limited knowledge of novel situations. The rigid nature of these policies frequently leads to suboptimal or unsafe behavior when encountering unfamiliar traffic conditions. Finally, although more sample-efficient than reinforcement learning, DNNIL methods still suffer from data inefficiency, requiring millions of state-action pairs for effective learning~\cite{fang2019survey,yu2023reinforcement}. These limitations call for further research aimed at addressing the transparency, generalizability, and data efficiency challenges of imitation learning.

To improve the transparency and interpretability of imitation learning, explainable AI (XAI) methods~\cite{saarela2024recent, dovsilovic2018explainable,nazat2024evaluating} often employ one of the following approaches: white-box (symbolic) models, explainable neural networks, or neurosymbolic frameworks. In white-box models, for example, a learning framework combining imitation learning and logical automata was proposed by~\citet{leech2019explainable} to represent problems as compact finite-state machines with human-interpretable logic states. Additionally,~\citet{bewley2020modelling} employed decision trees to interpret emulated policies using only inputs and outputs, while~\cite{zhang2021explainable} leveraged a hierarchical approach to ensure interpretability. On the other hand, pixel-wise CNN-based methods, which are often used in computer vision applications, capture high-level features using heatmaps and their implications~\cite{pan2020xgail}.

Furthermore, neurosymbolic learning methods—considered among the most cutting-edge—aim to combine the learning capabilities of DNNs with symbolic reasoning~\cite{sarker2021neuro,hitzler2022neuro}. Most neurosymbolic methods employ symbolic, logic-based reasoning to extract domain-specific logical rules using various rule-generation techniques~\cite{kimura2021neuro,zimmer2021differentiable}. As such, they are recognized as sample-efficient approaches that exhibit strong generalizability~\cite{keller2025neuro}.

Each of the three aforementioned approaches has its own advantages and limitations. Neurosymbolic frameworks like the Differentiable Logic Machine (DLM)~\cite{zimmer2021differentiable, song2022interpretable} integrate differentiable reasoning layers within deep neural architectures, enabling end-to-end training. However, they often require large-scale labeled data and may sacrifice full symbolic transparency due to the latent nature of their learned predicates. Pixel-wise CNN-based techniques are commonly applied in visual domains and also need large-scale datasets. In contrast, white-box models can provide explicit logical expressions behind learned behaviors using limited data. Yet, models based on finite-state machines and decision trees often struggle with scalability and problem complexity in challenging tasks.

As a white-box symbolic approach, search-based heuristic methods such as Inductive Logic Programming (ILP)~\cite{muggleton2012ilp, cropper2022inductive} have demonstrated the ability to efficiently extract abstract rules from a small number of examples when background knowledge is provided. Unlike finite-state machines, ILP-based approaches can scale better and handle more complex tasks by effectively searching for rules that satisfy the given examples. ILP can be employed to imitate human behavior using a limited number of examples; however, it has not yet been applied in the context of autonomous driving.

In this paper, we propose a novel rule-based imitation learning technique, called SIL, which is the first purely ILP-based imitation learning method in the context of autonomous driving. This method aims to generate explainable driving policies instead of black-box, non-interpretable ones by extracting symbolic first-order rules from human-labeled driving scenarios, using basic background knowledge provided by humans. It addresses the transparency and generalizability challenges associated with current neural network-based imitation learning methods. By extracting abstract logical relationships between states and actions in autonomous highway driving, SIL aims to deliver transparent, interpretable, and adaptive driving policies that enhance the safety and reliability of autonomous driving systems.

The main contributions of this paper are:

\begin{itemize}
    \item We propose SIL, a logic-based imitation learning method that extracts the underlying logical rules governing human drivers' actions in various scenarios. This approach enhances the transparency of learned driving policies by inducing human-readable and interpretable rules that capture essential aspects of safety, legality, and smoothness in driving behavior. Furthermore, SIL improves the generalizability of these policies, enabling AVs to handle diverse and challenging driving conditions.
    
    \item We compare SIL with state-of-the-art neural-network-based imitation learning methods using real-world HighD and NGSim datasets, demonstrating how a symbolic approach can outperform neural methods—even when trained on a small number of synthetic scenarios labeled as examples for each action.
\end{itemize}

The remainder of the paper is organized as follows. Section~\ref{sec:materialsmetot} introduces the prerequisites of the method and section~\ref{sec:methodology} describes the proposed approach in general. Section~\ref{sec:results} presents the simulation environment and experimental results, while Section~\ref{discussion} discusses and evaluates the outcomes. Finally, Section~\ref{conclusion} concludes the paper.

%%%%%%%%%%%%%%%%%%%%%%%%%%%%%%%%%%%%%%%%%%
\section{Background}\label{sec:materialsmetot}

%\section{Preliminaries}
This section presents the theoretical and methodological background relevant to the proposed approach.

\subsection{\textbf{First-Order Logic}}

First-order logic is a formalism that uses facts and rules to represent knowledge and perform logical inferences~\cite{yu2023reinforcement}. In this framework, a rule consists of a head and a body, written as \texttt{head :- body}, where \verb|:-| denotes the entailment operation~\cite{cropper2022inductive}. The \verb|head| represents an output predicate expressing a relationship between concepts, while the \verb|body| specifies the conditions under which the \verb|head| predicate holds.

Predicates are composed of a functor and a list of arguments, expressed as \verb|functor/n|, each of which has \verb|n| arguments that are either variables or constants. First-order logic rules typically follow the Horn clause structure, consisting of a single head literal and zero or more body literals:
\begin{equation}
\verb|H :- B_1, B_2, ..., B_n|,
\label{HBs}
\end{equation}
where \verb|B_i| ($i = 1, ..., n$) denotes the predicates in the \verb|body|, and commas (\verb|,|) indicate conjunctions ($\wedge$), consistent with Prolog syntax. The \verb|H| predicate serves as the conclusion of the rule. This structure implies that if all the \verb|B_i| predicates hold true conjunctively, then \verb|H| is true; otherwise, it is false.

First-order logic rules offer a robust mechanism for expressing complex relationships and performing logical reasoning based on structured knowledge.

\subsection{\textbf{Inductive Logic Programming}}

Inductive Logic Programming (ILP) is a machine learning paradigm that combines first-order logic with inductive reasoning to learn symbolic rules from examples, including a set of positive examples and, optionally, negative examples~\cite{cropper2022inductive}. These examples are typically represented as tuples of ground literals, where a ground literal is an atom in which all variables are instantiated as constants.

An ILP setup involves three key components: the language bias set $\mathcal{B}$, the background knowledge set $\mathcal{BK}$, and a set of examples $\mathcal{E}$. The set $\mathcal{B}$ defines the hypothesis space—that is, the rules the ILP system considers during learning. It includes a desired head predicate along with multiple candidate body predicates and guides the search for rules that could explain the given examples. The set $\mathcal{BK}$, which is domain-specific, provides additional contextual information, such as known relationships, constraints, or regularities, that can assist in inferring new rules. The set of examples $\mathcal{E}$ includes both positive examples $\mathcal{E}^+$ and negative examples $\mathcal{E}^-$, representing the observed instances from which the ILP system learns.

The primary goal of an ILP system is to discover hypotheses $\mathcal{H}$ that explain $\mathcal{E}^+$ while avoiding $\mathcal{E}^-$. The system begins by scanning $\mathcal{E}^+$ and attempting to match each example with every $b \in \mathcal{B}$. If a candidate $b$ covers a positive example according to $\mathcal{BK}$, it is added to the list of learned hypotheses. In the next step, any hypothesis that also covers a negative example from $\mathcal{E}^-$ is removed. The algorithm then refines the remaining hypotheses using $\mathcal{E}^+$ and $\mathcal{BK}$. For each hypothesis that covers a positive example $e^+ \in \mathcal{E}^+$, a more specific rule is generated by adding new literals to extend its coverage. Algorithm~\ref{algo:ilp} presents the pseudo-code of the ILP process.

ILP leverages the strengths of first-order logic and inductive reasoning to learn interpretable and generalizable symbolic rules from a small set of examples $\mathcal{E}$—even from a single example~\cite{cropper2022inductive}. By iteratively refining candidate rules under the guidance of $\mathcal{E}^+$ and $\mathcal{BK}$, ILP provides a principled framework for learning explainable policies in complex domains such as autonomous driving.
\begin{algorithm}[tb]
    \caption{Inductive Logic Programming (ILP)}
    \label{algo:ilp}
    \begin{algorithmic}[1]
        \REQUIRE $\mathcal{B}$ (bias set), $\mathcal{BK}$ (background knowledge), $\mathcal{E}^+$ (positive examples), $\mathcal{E}^-$ (negative examples)
        \ENSURE $\mathcal{H}$ (final set of learned hypotheses)

        \STATE Initialize $\mathcal{H}_{\text{old}} \leftarrow [\;]$ \hfill $\vartriangleright$ \textit{Start with an empty hypothesis set}

        \FOR{each $e^+ \in \mathcal{E}^+$}
            \FOR{each $b \in \mathcal{B}$}
                \IF{$b$ covers $e^+$ under $\mathcal{BK}$}
                    \STATE Add $b$ to $\mathcal{H}_{\text{old}}$ \hfill $\vartriangleright$ \textit{Retain $b$ as a candidate hypothesis}
                \ENDIF
            \ENDFOR
        \ENDFOR

        \FOR{each $e^- \in \mathcal{E}^-$}
            \FOR{each $h \in \mathcal{H}_{\text{old}}$}
                \IF{$h$ covers $e^-$ under $\mathcal{BK}$}
                    \STATE Drop $h$ from $\mathcal{H}_{\text{old}}$ \hfill $\vartriangleright$ \textit{Eliminate hypotheses that incorrectly cover negative examples}
                \ENDIF
            \ENDFOR
        \ENDFOR

        \STATE Initialize $\mathcal{H}_{\text{new}} \leftarrow [\;]$ \hfill $\vartriangleright$ \textit{Prepare to refine remaining hypotheses}

        \FOR{each $e^+ \in \mathcal{E}^+$}
            \FOR{each $h \in \mathcal{H}_{\text{old}}$}
                \IF{$h$ covers $e^+$ under $\mathcal{BK}$}
                    \STATE $h_r \leftarrow$ refinement of $h$ using $e^+$ \hfill $\vartriangleright$ \textit{Add new literals to make $h$ more specific}
                    \STATE Add $h_r$ to $\mathcal{H}_{\text{new}}$
                \ENDIF
            \ENDFOR
        \ENDFOR

        \STATE $\mathcal{H} \leftarrow \mathcal{H}_{\text{new}}$ \hfill $\vartriangleright$ \textit{Final set of refined, valid hypotheses}
    \end{algorithmic}
\end{algorithm}

\subsection{\textbf{Imitation Learning}}

Imitation learning is a widely used technique in machine learning and robotics that enables an agent to acquire complex behaviors by imitating expert demonstrations. In deep neural network-based imitation learning (DNNIL), the agent uses deep neural networks (DNNs) to learn from a dataset $\mathcal{D}$ consisting of expert demonstrations, represented as state-action pairs: $\{(s_1, a_1), (s_2, a_2), \ldots, (s_T, a_T)\}$. Here, $s_t$ denotes the state of the environment at time step $t$, and $a_t$ is the action taken by the expert in that state. The objective of DNNIL is to learn a policy $\pi_\theta(a|s)$, parameterized by $\theta$, that maps states to actions in a manner consistent with expert behavior.

A common approach to training the imitation policy is through supervised learning. The learning process involves minimizing the discrepancy between the actions predicted by the learned policy and the actions taken by the expert. This discrepancy is typically measured using a mean squared error (MSE) loss function, as shown in Eq.~\ref{loss}. The aim is to identify the optimal parameters $\theta^*$ that minimize this loss over the dataset $\mathcal{D}$:
\begin{equation}
\theta^* = \arg\min_\theta \sum_{(s, a) \in \mathcal{D}} (\pi_\theta(a|s) - a)^2 
\label{loss}
\end{equation}

DNNIL has demonstrated significant success in various domains, including autonomous driving. By learning directly from expert drivers’ demonstrations, DNNIL agents are capable of handling complex traffic scenarios. However, one of the major challenges associated with DNNIL is its lack of interpretability. Since policies are encoded within deep networks, understanding the decision-making process becomes difficult. Moreover, DNNIL tends to perform poorly outside the distribution of the training data, resulting in limited generalization to unseen situations.

\section{Methodology: Symbolic Imitation Learning}\label{sec:methodology}

\begin{figure}[H]
\begin{adjustwidth}{-\extralength}{0cm}
\centering
\includegraphics[width=1\linewidth, trim={1cm 20cm 1cm 5cm}, clip]{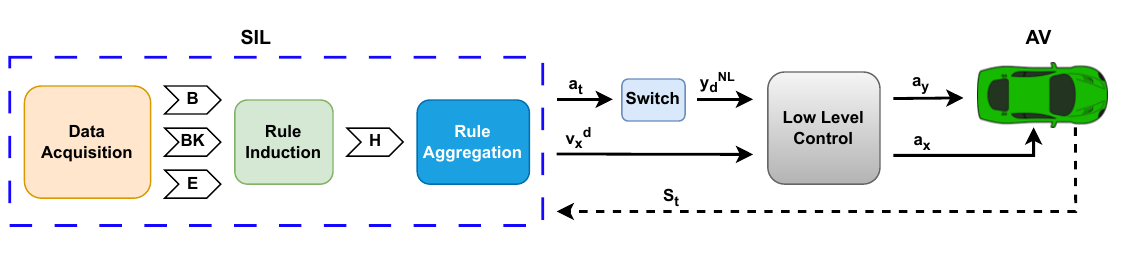}
\caption{
Symbolic imitation learning (SIL) framework for autonomous driving decision-making. 
In the SIL module, \textbf{B} denotes background facts from the environment, \textbf{BK} is background knowledge, and \textbf{E} represents training examples. 
Rule induction produces a set of rules \textbf{H}, from which \texttt{h1}–\texttt{h5} filter fatal actions and \texttt{h6}–\texttt{h7} rank the safe candidates to select the optimal action among: lane keeping (LK), left lane change (LLC), and right lane change (RLC). 
The chosen action $a_t$ is mapped to a Next Lane (NL) desired lateral position $y_d^{NL}$ via a switch, and a PID controller generates the lateral acceleration $a_y$, combined with the longitudinal acceleration $a_x$ to control the AV.
}

%\caption{Decision-making using symbolic imitation learning framework in autonomous driving.}
\label{sil}
\end{adjustwidth}
\end{figure}

The Symbolic Imitation Learning (SIL) framework introduces a novel approach that leverages ILP to extract symbolic policies from human-generated background knowledge. The core objective of this framework is to replicate human behaviors by uncovering explicit rules that govern complex actions demonstrated by humans. As illustrated in Fig.~\ref{sil}, SIL comprises three main components: knowledge acquisition, rule induction, and rule aggregation.

In the knowledge acquisition phase, essential inputs are provided based on prior knowledge about the environment. These include the language bias set $\mathcal{B}$, background knowledge $\mathcal{BK}$, and the set of examples $\mathcal{E}$ (refer to Algorithm~\ref{algo:ilp}). These components enable the ILP system to induce a single rule during the rule induction phase. To construct a complete policy, this process is repeated iteratively for all required rules, progressively assembling them into the hypothesis set $\mathcal{H}$. The final rule aggregation component then utilizes and refines these induced rules to infer the desired actions. Since these logical rules are derived directly from human demonstrations, the resulting actions are inherently interpretable and human-like. By capitalizing on ILP's ability to infer symbolic rules from structured examples, SIL effectively captures nuanced human behavior—an area where conventional DNNIL methods often face limitations.

One of the key advantages of SIL is its sample efficiency. It can generate a coherent and interpretable set of rules from a relatively small number of expert demonstrations. These rules not only support human-understandable decision-making but also enhance the system's generalizability. The explicit, symbolic nature of the learned policies enables SIL-based agents to better adapt to previously unseen scenarios, outperforming black-box policies that typically lack both transparency and adaptability.
\begin{comment}
\begin{figure*}[ht]
	\vskip 0.2in
	\begin{center}
		\centerline{\includegraphics[width=0.9\textwidth]{SIL2.png}}
		\caption{Decision-making using symbolic imitation learning framework in autonomous driving}
		\label{sil}
	\end{center}
	\vskip -0.2in
\end{figure*}
\end{comment}

% \begin{figure}[H]
% \begin{adjustwidth}{-\extralength}{0cm}
% \centering
% \includegraphics[width=\linewidth]{symbolic_imitation.drawio.png}
% \caption{Decision-making using symbolic imitation learning framework in autonomous driving.}
% \label{sil}
% \end{adjustwidth}
% \end{figure}

\textbf{Symbolic Imitation Learning in Autonomous Driving:}
As a use-case scenario, SIL aims to derive unknown rules for autonomous highway driving. The primary objective of employing SIL in this context is to extract driving rules from human-derived background knowledge and use them to emulate human-like behavior. These behaviors include lane changes and adjustments to the AV's longitudinal velocity, all of which are essential for achieving safe, efficient, and smooth driving. During lane-change decisions, the AV can execute one of three discrete actions: LK, RLC, or LLC.

\begin{figure*}[h]
\begin{adjustwidth}{-\extralength}{0cm}
\centering
\includegraphics[width=0.8\linewidth, trim={1cm 12cm 1cm 10cm}, clip]{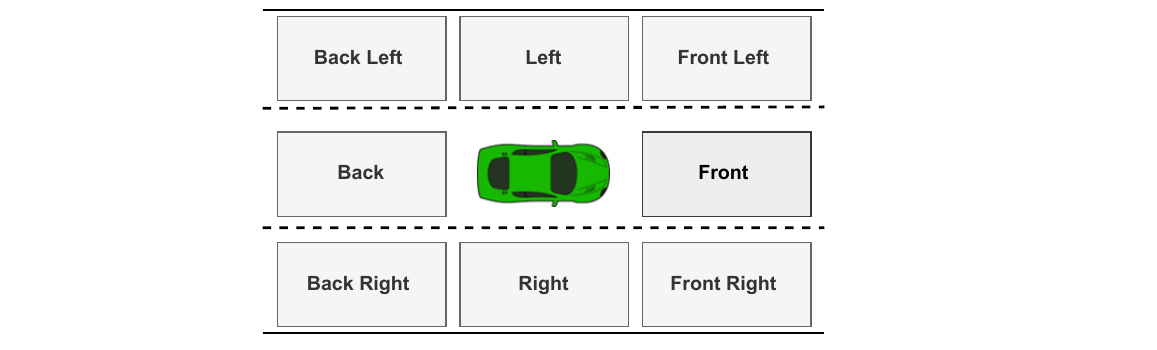}
\caption{
Sector-based representation of the autonomous vehicle (AV) on a three-lane highway. 
The green vehicle denotes the AV, and the surrounding eight sectors (Front, Front Left, Front Right, Left, Right, Back, Back Left, Back Right) define its state space. 
Each sector encodes whether it is occupied or not.%, as well as the relative distance and velocity of any vehicle present.
}

\label{sectors}
\end{adjustwidth}
\end{figure*}

\textbf{Background:} Human drivers frequently adjust their lane position and velocity based on the positions and speeds of nearby vehicles, also referred to as target vehicles (TVs). Motivated by this principle, the proposed approach incorporates the relative positions and velocities of TVs surrounding the AV to support informed decision-making regarding lane changes and speed adjustments. To formalize this, the area surrounding the AV is partitioned into eight distinct \verb|sector|s: \verb|front|, \verb|frontRight|, \verb|right|, \verb|backRight|, \verb|back|, \verb|backLeft|, \verb|left|, and \verb|frontLeft|, as shown in Figure~\ref{sectors}. Each sector may either be occupied by a vehicle—indicated by the predicate \verb|sector_isBusy| set to \verb|true|—or unoccupied, in which case the predicate is set to \verb|false|. For example, \verb|right_isBusy| is \verb|true| if there is a vehicle in the \verb|right| sector; otherwise, it is set to \verb|false|. Table~\ref{tab:predicate-definitions} in the Appendix shows all the necessary predicates used in this research with their definitions. Additional details on predicate definitions are provided in~\cite{sharifi2025toward}.

To capture the relative longitudinal velocities of TVs, we compute the difference between each TV's velocity and that of the AV, then assign predicates accordingly. Each sector is associated with three predicates reflecting this velocity difference. If the relative velocity is greater (or smaller) than a predefined threshold $\eta_v$, the predicate \verb|bigger| (or \verb|lower|) is used. Threshold $\eta_v$ ensures that velocity differences are significant rather than negligible, and is set to 5~km/h in this context. When the absolute value of the relative velocity is within the threshold range, it is considered \verb|equal|. For example, the predicates \verb|frontVel_isBigger|, \verb|frontVel_isEqual|, and \verb|frontVel_isLower| describe the relative velocity of the TV in the \verb|front| sector with respect to the AV.

In addition, AVs—like human drivers—must remain within valid road sections (i.e., highway lanes) and avoid entering off-road or restricted areas. To represent this contextual constraint, each sector is also labeled as either valid (\verb|sector_isValid| is set to \verb|true|) or invalid (\verb|sector_isValid| is set to \verb|false|).

\subsection{\textbf{Knowledge Acquisition}} \label{data-acq}

Autonomous driving systems require a variety of rules to perform effectively under diverse conditions, and each rule must be learned under a consistent setting with appropriate examples. Therefore, to extract distinct rules, it is essential to define unique configurations and supply corresponding examples for each one. We begin by specifying the scope of each desired rule through the definition of its head predicate and a set of candidate body predicates, all encapsulated within a specific bias set $\mathcal{B}$. The selection of candidate body predicates depends on their potential impact on the accuracy of the head predicate. Consequently, body predicates that have no effect on the head are discarded, reducing the dimensionality of $\mathcal{B}$ and avoiding excessive computation during the learning process.

To identify the rules, we initially categorize the actions into four unique sets: \textit{fatal}, \textit{risky}, and \textit{efficient} lane-change actions, along with \textit{smooth} longitudinal velocities. Fatal actions are the actions leading to serious accidents or crashes, while risky actions are mildly dangerous actions that might lead to law violations. Efficient actions are associated with the actions leading to not only smooth lane changes but also higher level of safety. These categories can be interdependent.

For each category of actions, we provide human-labeled datasets consisting of scenarios with an ego vehicle surrounded by a varying number of intruders. In each of the scenario, taking a certain action is either true or false. For example, in the fatal lane change dataset, if there is an intruder on the left, taking the left action is considered as fatal. These true/false labels help to define positive and negative examples suitable for ILP systems. Table~\ref{rules} indicates the number of positive and negative examples in each action category with the number related body predicates to the corresponding action. In general, the goal is to extract rules associated with the aforementioned categories to ensure safe and efficient lateral lane changes and longitudinal velocity control.

In each category, the corresponding rule has a unique head predicate defined in $\mathcal{B}$ using the \verb|head_pred/1| declaration, and a corresponding set of candidate body predicates. These candidate body predicates include all literals that may influence the head predicate. Then, the ILP system searches for optimal body predicates that satisfy all positive examples while simultaneously rejecting negative examples. To enable this, we define multiple possible scenarios using background knowledge and assign a unique identifier to each scenario. Based on expert understanding of the intended action in each scenario, scenarios are labeled as either positive examples using the \verb|pos/1| predicate or negative examples using the \verb|neg/1| predicate. This process is repeated for each rule to generate sufficient knowledge-driven data for inducing previously unknown logical rules. As such, each rule induction task requires a specific configuration of $\mathcal{B}$, background knowledge $\mathcal{BK}$, and training examples $\mathcal{E}$ to support effective learning.

\begin{table}[htbp]
    %\caption{Details of the rules extracted during the knowledge acquisition and rule induction processes.}
    \caption{
Summary of the datasets and the rules extracted during the knowledge acquisition and rule induction phases of the SIL framework.
The table reports the rule head, the number of predicates in the bias set ($N_{\text{BP}}$), the counts of positive ($N_{\mathcal{E}^+}$) and negative ($N_{\mathcal{E}^-}$) examples used for induction, the hypothesis label, the achieved accuracy, and the induction time ($T_i$). 
Abbreviations: $N_{\text{BP}}$—number of predicates in bias set $\mathcal{B}$; $N_{\mathcal{E}^+}$—number of positive examples; $N_{\mathcal{E}^-}$—number of negative examples; $T_i$—induction time.
}
    \label{rules}
    \begin{adjustwidth}{-\extralength}{0cm}
        \begin{tabularx}{\fulllength}{>{\raggedright\arraybackslash}p{3cm} l l c c c c c c c}
            \toprule
            \textbf{Dataset Category} & \textbf{Action} & \textbf{Head Predicate} & $\textbf{N}_{\textbf{BP}}$ & $\textbf{N}_{\mathcal{E}^+}$ & $\textbf{N}_{\mathcal{E}^-}$ & \textbf{Hypothesis} & \textbf{Accuracy} & \textbf{$\textbf{T}_\textbf{i}$ (sec)} \\
            \midrule
            \multirow{2}{*}{\makecell[l]{\textbf{Fatal} \\  \textbf{Lane Changing} }} 
                & RLC & \texttt{rlc\_isFatal}    & 38 & 768 & 256 & \texttt{h1}  & 1.00 & 0.365 \\
                & LLC & \texttt{llc\_isFatal}    & 38 & 768 & 256 & \texttt{h2}  & 1.00 & 0.345 \\
            \midrule
            \multirow{3}{*}{\makecell[l]{\textbf{Risky} \\ \textbf{Lane Changing} }} 
                & LK  & \texttt{lk\_isRisky}  & 43 & 87  & 937 & \texttt{h3}  & 1.00 & 0.762 \\
                & RLC & \texttt{rlc\_isRisky} & 38 & 320 & 704 & \texttt{h4}  & 1.00 & 0.416 \\
                & LLC & \texttt{llc\_isRisky} & 38 & 308 & 717 & \texttt{h5}  & 1.00 & 1.007 \\
            \midrule
            \multirow{2}{*}{\makecell[l]{\textbf{Efficient} \\ \textbf{Lane Changing} }} 
                & RLC & \texttt{rlc\_isBetter}    & 25 & 32  & 992 & \texttt{h6}  & 1.00 & 4.108 \\
                & LLC & \texttt{llc\_isBetter}    & 25 & 128 & 896 & \texttt{h7}  & 1.00 & 0.892 \\
            \midrule
            \multirow{3}{*}{\makecell[l]{\textbf{Smooth} \\ \textbf{Longitudinal} \\ \textbf{Velocity} }} 
                & Catch-up  & \texttt{reachDesiredSpeed} & 43 & 512 & 512 & \texttt{h8} & 1.00 & 0.336 \\
                & Follow-up & \texttt{reachFrontSpeed}   & 43 & 253 & 771 & \texttt{h9} & 1.00 & 0.703 \\
                & Brake     & \texttt{brake}             & 43 & 253 & 771 & \texttt{h10} & 1.00 & 0.326 \\
            \bottomrule
        \end{tabularx}
    \end{adjustwidth}
%\noindent{\footnotesize $N_{\text{BP}}$: Number of predicates in bias set $\mathcal{B}$; $N_{\mathcal{E}^+}$: Number of positive examples; $N_{\mathcal{E}^-}$: Number of negative examples; $T_i$: Induction time.}
\end{table}

The knowledge acquisition process generates a variety of real-world scenarios representing various driving scenarios labeled by a real human driver. Each state consists of an AV surrounded by eight spatial sections, each of which is categorized as either occupied by a TV or vacant. Accordingly, we include eight \verb|sector_isBusy| literals in the candidate body predicates—one for each \verb|sector| surrounding the AV.

Additionally, for each state, we incorporate the relative velocity of every TV with respect to the AV. This results in 24 relative velocity literals—three per \verb|sector|—being added to the candidate body predicates. To ensure legal compliance in driving decisions, we also consider the validity of the \verb|right| and \verb|left| sections, adding two corresponding literals to represent whether these areas are drivable. Table~\ref{rules} summarizes the total number of candidate body predicates associated with each head predicate.

Once the states are defined, we assign positive or negative labels to the head predicates of the target rules based on expert driving knowledge and the behavioral outcome expected in each scenario. For example, to guide the ILP system in deriving fatal RLC rules, we label states in which taking the RLC action is considered fatal as positive examples; otherwise, such states are labeled as negative examples.

The number of predicates in each bias set $\mathcal{B}$, along with the counts of positive and negative examples, are denoted by $N_{\text{BP}}$, $N_{\mathcal{E}^+}$, and $N_{\mathcal{E}^-}$, respectively. These parameters significantly influence the accuracy and robustness of the induced rules. While including negative examples ($\mathcal{E}^-$) is optional, their presence improves the generalizability and resilience of the rules across a wider range of environmental states. As indicated in Table~\ref{rules}, distinct values of $N_{\mathcal{E}^+}$ and $N_{\mathcal{E}^-}$ are defined for each rule.
\footnote{All knowledge-based datasets are available at \href{https://github.com/CAV-Research-Lab/Symbolic-Imitation-Learning/tree/main/SIL/data}{github.com/CAV-Research-Lab/Symbolic-Imitation-Learning}}

\subsection{\textbf{Rule Induction}}

After obtaining a sufficient amount of knowledge-driven data for each target rule, we proceed to the rule induction stage of the SIL framework. This stage focuses on learning interpretable rules tailored to autonomous highway driving. The rule extraction process is carried out using Popper, a state-of-the-art ILP system that integrates Answer Set Programming\footnote{ASP is a declarative programming paradigm well-suited for solving combinatorial and knowledge-intensive problems by encoding them as logic-based rules and constraints.} (ASP)~\cite{lifschitz2019answer, corapi2011inductive} with Prolog to enhance learning efficiency and accuracy.

One of Popper’s main advantages is its ability to learn from failures (LFF). This is achieved through a three-stage process: generating candidate rules by exploring the hypothesis space via ASP (generate stage), evaluating those candidates against the positive examples $\mathcal{E}^+$ and background knowledge $\mathcal{BK}$ using Prolog (test stage), and pruning the hypothesis space based on failed hypotheses (constrain stage)~\cite{cropper2021learning}.

Given the dynamic nature of decision-making in autonomous driving, a wide range of rules can be induced for different tasks. However, in this work, we focus on extracting only the essential general rules required for highway driving, specifically in the categories of \textit{safety}, \textit{efficiency}, and \textit{smoothness}.

\subsubsection{\textbf{Safe Lane Changing}}
%This section aims to find unsafe lane changing actions from human-labeled scenarios, helping to eliminate them from the action space, which leads to a safe action space. To find the unsafe actions, we consider two datasets with the following types of unsafe actions: \textit{fatal} and \textit{risky} lane changing actions. The RLC and LLC actions may fall into either category depending on the surrounding traffic conditions. In contrast, LK is generally assumed to be safe, except in specific situations where it may pose a potential danger.

This section aims to identify unsafe lane-changing actions from human-labeled scenarios, thereby eliminating them from the action space and ensuring that only safe actions remain. To this end, we consider two datasets containing the following types of unsafe actions: \textit{fatal} and \textit{risky} lane-changing actions. The RLC and LLC actions may fall into either category depending on the surrounding traffic conditions. In contrast, LK is generally assumed to be safe, except in specific situations where it may pose potential danger.

\begin{figure*}[t]

    \centering
    \subfloat[]{\includegraphics[width=0.3\textwidth, height=0.20\textwidth]{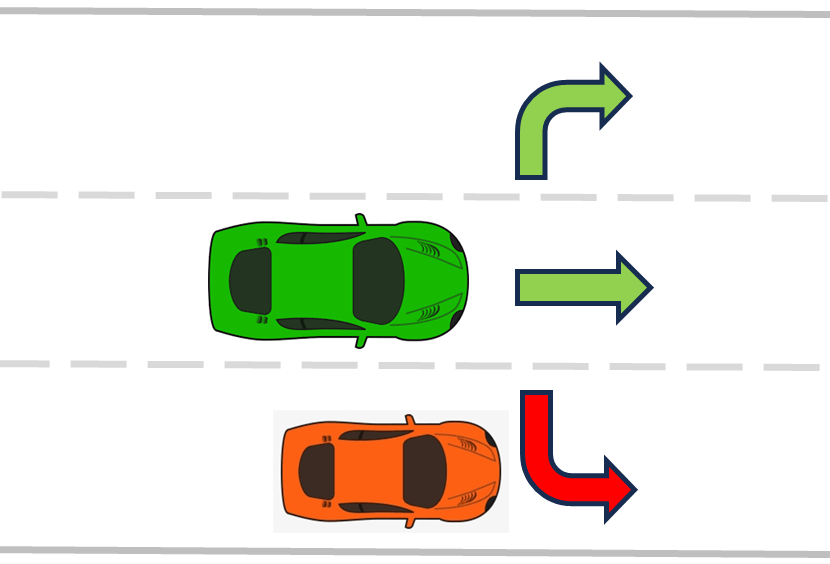}}\hfill
    \subfloat[]{\includegraphics[width=0.3\textwidth, height=0.20\textwidth]{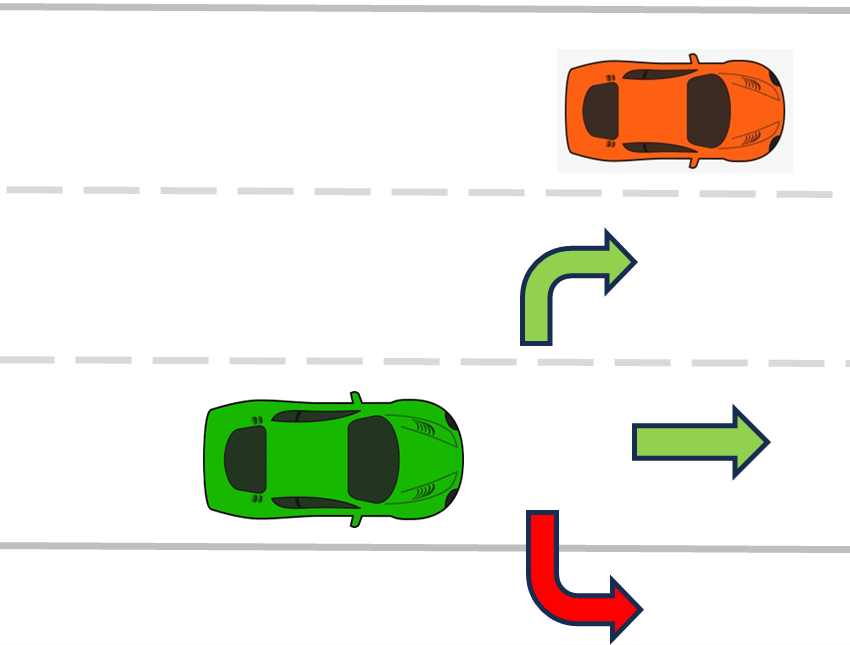}}\hfill
    \subfloat[]{\includegraphics[width=0.3\textwidth, height=0.20\textwidth]{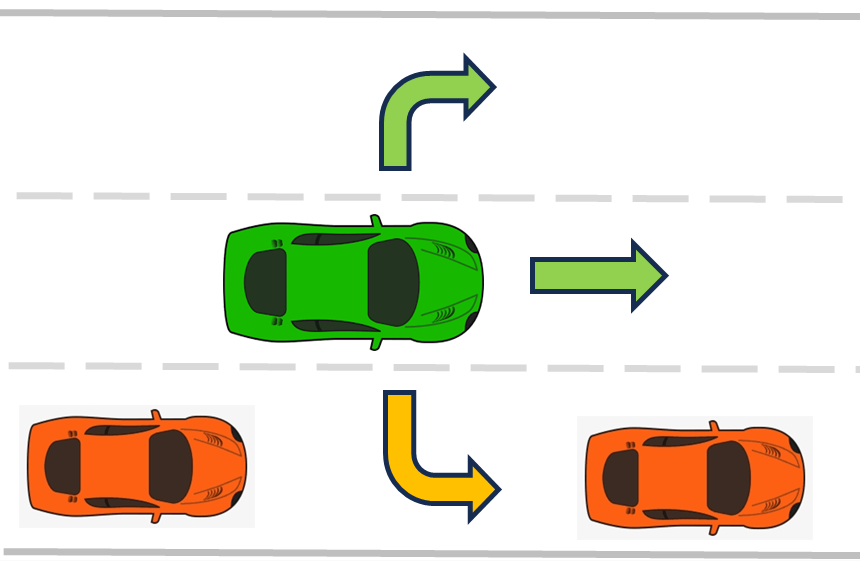}}\par\vspace{0.5em}
    \subfloat[]{\includegraphics[width=0.3\textwidth, height=0.20\textwidth]{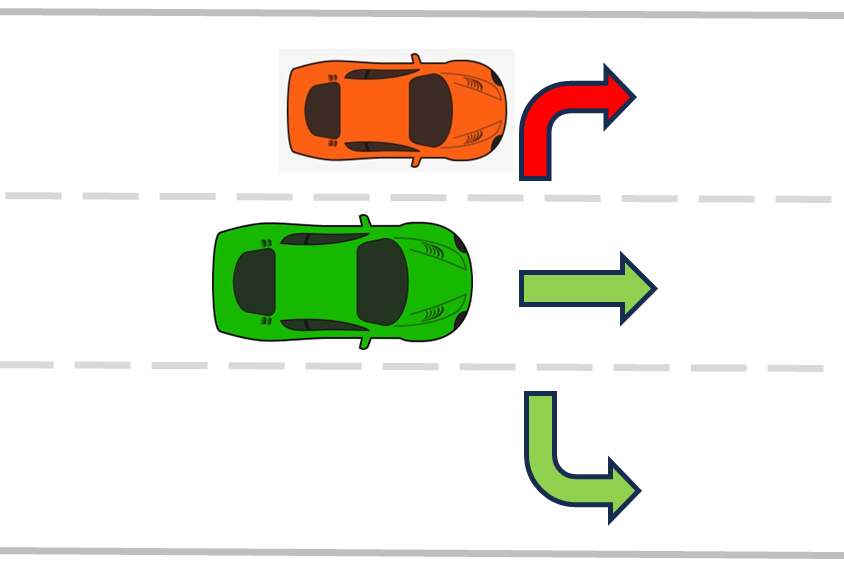}}\hfill
    \subfloat[]{\includegraphics[width=0.3\textwidth, height=0.20\textwidth]{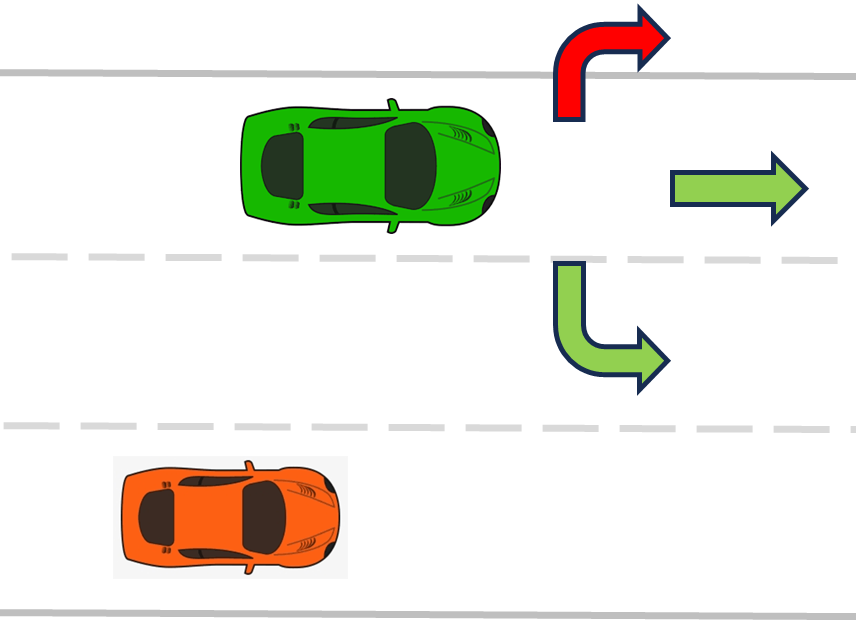}}\hfill
    \subfloat[]{\includegraphics[width=0.3\textwidth, height=0.20\textwidth]{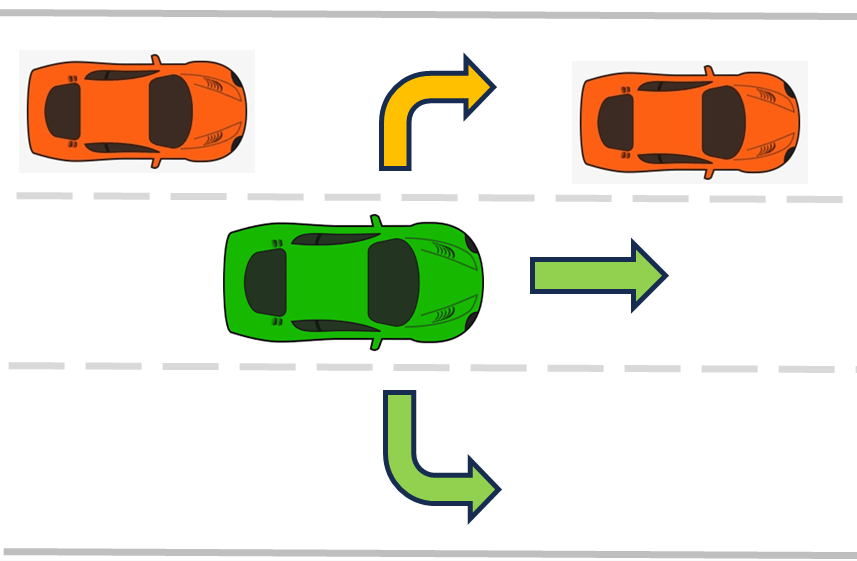}}
    \caption{Samples of positive and negative examples for fatal and risky lane change rules. For the top-left (a) and top-middle (b) scenarios, \texttt{pos(rlc\_isFatal)} and \texttt{neg(llc\_isFatal)}, and also for the bottom-left (d) and bottom-middle (e) scenarios, \texttt{neg(rlc\_isFatal)} and \texttt{pos(llc\_isFatal)} hold true. For the top-right (c) scenario, \texttt{pos(rlc\_isRisky)} and \texttt{neg(llc\_isRisky)}, and also for the bottom-right (f) scenario, \texttt{neg(rlc\_isRisky)} and \texttt{pos(llc\_isRisky)} hold true.}
    \label{scenarios}
\end{figure*}

\textbf{Fatal Lane Changing:} The objective here is to uncover previously unknown rules that characterize situations in which executing RLC or LLC would be fatal and could lead to collisions. For instance, four such scenarios are illustrated in Fig.~\ref{scenarios}—the top-left and top-middle subfigures correspond to fatal RLC examples, while the bottom-left and bottom-middle represent fatal LLC examples. Using ILP, we derive two rules: one for fatal RLC (\verb|h1|) and another for fatal LLC (\verb|h2|). For example, rule \verb|h1| specifies when taking the RLC action is considered fatal:

\begin{rulebox}{Rule h1: Fatal RLC}
rlc_isFatal:- right_isBusy; not(right_isValid).
\end{rulebox}

Here, the comma \verb|(,)| denotes conjunction, the semicolon \verb|(;)| represents disjunction, and \verb|not(.)| indicates negation, following first-order logic notation. Similarly, rule \verb|h2| identifies conditions under which executing LLC is fatal:
\begin{rulebox}{Rule h2: Fatal LLC}
llc_isFatal:- left_isBusy; not(left_isValid).
\end{rulebox}

In summary, if a TV is present in the \verb|right| (or \verb|left|) section, or if that section is marked as invalid, performing an RLC (or LLC) action is deemed unsafe due to the high risk of collision. Therefore, the AV should avoid these actions in such states.

\textbf{Risky Lane Changing:} These actions refer to situations in which the AV selects maneuvers that, while not perilous, may still pose a risk to passenger safety. For instance, as illustrated in the top-right image of Fig.~\ref{scenarios}, consider a scenario where a vehicle occupies the \verb|backRight| (or \verb|backLeft|) section of the AV, and its velocity is higher than that of the AV. Although executing an RLC (or LLC) maneuver may be legally or physically feasible, doing so could lead to a collision with the approaching vehicle in the adjacent section. In such cases, the action is not fatal but is considered hazardous and should ideally be avoided.

Using the Popper ILP system, we derive three rules that characterize risky actions. The first rule (\verb|h3|) specifies that if there is a vehicle in the \verb|backRight| section with a higher velocity than the AV, or if the \verb|frontRight| section is occupied by a slower-moving vehicle, then performing an RLC maneuver is deemed risky:
\begin{rulebox}{Rule h3: Risky RLC}
rlc_isRisky:-
  backRight_isBusy, backRightVel_isBigger; 
  frontRight_isBusy, frontRightVel_isLower.
\end{rulebox}

Similarly, rule \verb|h4| captures the conditions under which an LLC maneuver is risky:
\begin{rulebox}{Rule h4: Risky LLC}
llc_isRisky:-
  backLeft_isBusy, backLeftVel_isBigger;
  frontLeft_isBusy, frontLeftVel_isLower.
\end{rulebox}

Finally, rule \verb|h5| addresses cases where the LK action may pose danger. It states that if a TV is present in the \verb|back| section, the distance to the AV is unsafe, and the TV is approaching at a higher velocity, then remaining in the current lane becomes hazardous:
\begin{rulebox}{Rule h5: Risky LK}
lk_isRisky:- back_isBusy, not(backDist_isSafe), backVel_isBigger.
\end{rulebox}

This rule highlights that even lane keeping—typically considered the safest option—can become endangered when the AV is being rapidly approached from behind without adequate spacing, thereby increasing the risk of rear-end collisions.

\subsubsection{\textbf{Efficient Lane Changing}}

While safety rules identify lane-change actions that are either fatal or risky lane changes in specific traffic conditions, they do not provide guidance on which action is more efficient when none of the options pose safety risks. To address this, we introduce a prioritization scheme that helps the AV make time-efficient decisions.

In general, minimizing driving time requires the AV to change lanes when appropriate. However, unnecessary or abrupt lane changes can compromise passenger comfort and health. Therefore, the default priority is to maintain the current lane unless a lane change is deemed necessary. When the AV needs to overtake a vehicle in the \verb|front| section, we introduce a secondary prioritization: if both RLC and LLC are viable, the AV should prefer LLC, as the left lane typically supports higher traffic speeds. RLC should be considered only when LLC is not feasible. This rule-based preference allows the AV to make more efficient decisions while preserving safety and comfort.

Based on this prioritization strategy, we label scenarios from the knowledge-driven dataset to reflect the more favorable action and use ILP to learn the corresponding rules. Rule \verb|h6| identifies a situation where LLC is preferable to RLC:
\begin{rulebox}{Rule h6: Efficient LLC}
llc_isBetter:- front_isBusy, not(left_isBusy), not(frontLeft_isBusy).
\end{rulebox}

This rule specifies that when the \verb|front| section is occupied and both the \verb|left| and \verb|frontLeft| sections are clear, the AV should prefer LLC over RLC. Conversely, rule \verb|h7| describes situations in which RLC becomes the preferred action:
\begin{rulebox}{Rule h7: Efficient RLC}
rlc_isBetter:-
  front_isBusy, left_isBusy,
  frontLeft_isBusy, not(right_isBusy), not(frontRight_isBusy).
\end{rulebox}

This rule applies when the \verb|front|, \verb|left|, and \verb|frontLeft| sections are all occupied, while the \verb|right| and \verb|frontRight| sections are available. In such cases, RLC is the more efficient option.

Importantly, the predicates \verb|rlc_isBetter| and \verb|llc_isBetter| are mutually exclusive by design. If one holds true, the other cannot, thereby avoiding conflicting recommendations during decision-making.

\subsubsection{\textbf{Smooth Longitudinal Velocity Control}} \label{acceleration rules}

To ensure passenger comfort and overall safety, an AV—like a human driver—must adjust its velocity smoothly. Human drivers typically rely on a limited set of intuitive actions to manage vehicle speed in a continuous manner. These include gradually accelerating to reach a desired cruising speed, adjusting speed to match that of a slower vehicle ahead (referred to as the \verb|front| TV), and decelerating in response to unsafe following distances. In critical situations, such as when the AV is too close to the \verb|front| TV while moving faster, the driver—or the AV—must apply the brake to prevent a collision. Based on these observed driving behaviors, the goal is to identify three fundamental rules that govern safe and smooth velocity adjustments.

\begin{comment}
\begin{figure}[ht]
	\begin{center}
		\centerline{\includegraphics[width=0.8\linewidth]{acceleration.png}}
		\caption{Acceleration phases: catch-up (for reaching the driver’s desired speed), follow-up (for adjusting to the front TV’s speed), and brake (for necessary deceleration).}
		\label{acc_phases}
	\end{center}
\end{figure}
\end{comment}

\begin{figure}[H]
\begin{adjustwidth}{-\extralength}{0cm}
\centering
\includegraphics[width=1\linewidth, trim={1cm 20.5cm 1cm 4.5cm}, clip]{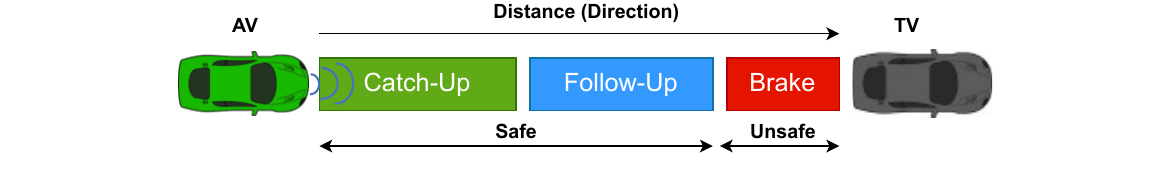}
\caption{
%Acceleration phases: catch-up (for reaching the driver’s desired speed), follow-up (for adjusting to the front TV’s speed), and brake (for necessary deceleration).
Acceleration phases for the autonomous vehicle (AV) as green vehicle and relative to the target vehicle (TV) as black vehicle. 
The \textit{catch-up} phase is used to reach the driver’s desired speed, the \textit{follow-up} phase adjusts the AV’s velocity to match the front TV’s speed while maintaining a safe distance, and the \textit{brake} phase is activated when the distance becomes unsafe, requiring deceleration. 
Safe and unsafe regions are indicated along the distance axis.
}

\label{acc_phases}
\end{adjustwidth}
\end{figure}

In the previous work~\cite{sharifi2025toward}, we introduced a rule-based method for longitudinal velocity control in autonomous vehicles. The results demonstrated that the AV could avoid collisions with the \verb|front| TV while ensuring smooth acceleration and deceleration transitions. This was achieved by integrating a low-level controller that eliminated discontinuities in the velocity commands. As shown in Eq.~\ref{accelerations}, the proposed approach models three distinct acceleration phases, inspired by typical lane-following behavior observed in human drivers, and aligns with methods proposed in~\cite{sun2021neuro}.

As illustrated in Fig.~\ref{acc_phases}, the \verb|catch-up| phase applies when the \verb|front| sector is unoccupied. In such cases, the AV has the flexibility to accelerate, decelerate, or maintain speed. To compute the appropriate acceleration, we define a desired longitudinal velocity term ($V^d_x$), which may differ across drivers or situations, and use it to determine the required acceleration to reach this target speed. For this task, $V^d_x$ is 110 (km/h).

When a vehicle is present in the \verb|front| section, human drivers typically adjust their speed to match that of the \verb|front| TV, thereby maintaining a safe distance and avoiding collisions. This behavior corresponds to the \verb|follow-up| phase, which applies when a TV occupies the \verb|front| sector and the longitudinal distance between the AV and the TV is considered safe. As shown in Eq.~\ref{accelerations}, the \verb|follow-up| phase computes the required acceleration that allows the AV to synchronize its velocity ($v_{x,AV}$) with that of the \verb|front| TV ($v_{x,TV}$), while gradually reducing the separation distance $D$.

Finally, the \verb|brake| phase applies in situations where the AV finds itself at an unsafe distance—specifically, when $D$ falls below a critical threshold $C$—and is traveling at a higher speed than the \verb|front| TV. In these emergency scenarios, braking is necessary to prevent a potential collision. The AV must decelerate promptly until a safe following distance is re-established. This phase is illustrated in Fig.~\ref{acc_phases} and formalized as the third case in the following equation:
\begin{equation}
a_x =
\begin{cases}
    \dfrac{V^d_x - v_{x,AV}}{\Delta t} & \quad \verb|catch-up|, \\[6pt]
    \dfrac{v^2_{x,TV} - v^2_{x,AV}}{2(D - C)} & \quad \verb|follow-up|, \\[6pt]
    \dfrac{-v^2_{x,AV}}{2D} & \quad \verb|brake|,
\end{cases}
\label{accelerations}
\end{equation}
where $D$ is the distance between the AV and the front TV, and $C$ is the critical braking distance which is considered 15 (m) in this use-case. The time step $\Delta t$ is 0.04 (sec).

As previously stated, the objective is to define logical rules that correspond to the acceleration conditions described in Eq.~\ref{accelerations}. However, existing ILP systems are not designed to reason directly over continuous numerical values (except for binary values such as 0 and 1), making it impractical to extract the complete equations symbolically. Instead, we identify which of the three control phases—\verb|catch-up|, \verb|follow-up|, or \verb|brake|—is applicable in a given state. Once the appropriate phase is determined, the corresponding acceleration is computed analytically.

To induce rules for each control phase, we construct training examples by evaluating the relative distance between the AV and the \verb|front| TV. Based on this evaluation, each state is labeled as a positive or negative example for the corresponding rule, as detailed in Table~\ref{rules}. This approach enables Popper to learn one symbolic rule per phase. For instance, the following rule specifies that when the \verb|front| section is unoccupied, the AV should enter the \verb|catch-up| phase to reach its desired speed:
\begin{rulebox}{Rule h8: Catch-up}
reachDesiredSpeed:-not(front_isBusy).
\end{rulebox}
The second rule addresses the \verb|follow-up| phase. It states that when the \verb|front| section is occupied and the distance between the AV and the \verb|front| TV is safe, the AV should adjust its acceleration to match the speed of the vehicle ahead:

\begin{rulebox}{Rule h9: Follow-up}
reachFrontSpeed:- front_isBusy, frontDist_isSafe.
\end{rulebox}

Finally, the third rule pertains to the \verb|brake| phase. It specifies that when a vehicle is present in the \verb|front| section, the relative distance is unsafe, and the AV is moving faster than the \verb|front| TV, the AV should initiate braking to prevent a collision:

\begin{rulebox}{Rule h10: Brake (deceleration)}
brake:- front_isBusy, frontVel_isLower, not(frontDist_isSafe).
\end{rulebox}

As summarized in Table~\ref{rules}, we evaluate the quality of each learned rule using an \textit{accuracy} metric, defined as the average of \textit{precision} and \textit{recall} over the corresponding positive ($\mathcal{E}^+$) and negative ($\mathcal{E}^-$) examples. According to the results, all induced rules achieve perfect accuracy (1.00), indicating that all the extracted rules successfully covered all considered scenarios and predict positive and negative examples within their respective domains. Since there is no mislabeled data in the datasets, it is not beyond expectation that an ILP-based system can find the rules satisfying all positive and negative examples.

In addition to accuracy, we also report the rule induction time ($T_i$) for each rule in Table~\ref{rules}. The values show that $T_i$ is consistently low across all rules, demonstrating that the SIL framework enables fast and efficient rule induction. This efficiency can be attributed to Popper’s structured approach, which combines symbolic reasoning with effective pruning strategies to learn concise and accurate rules.

To evaluate the robustness of this method, we intentionally flip the data labels randomly to observe how Popper manages the noise. Varying percentages of noise from 2\% to 20\% are considered for all the datasets. As shown in Table~\ref{tab:noisy_results} (see Appendix), increasing the noise percentage leads to a significant decrease in both precision and recall. Moreover, the induction time $T_i$ also increases.

\subsection{\textbf{Rule Aggregation}}

Having extracted the rules \verb|h1| to \verb|h10| using Popper, these interpretable rules are further processed to generate coherent, human-like decisions for lane changes and velocity adjustments. This post-processing is performed by the rule aggregation component, which integrates the induced rules into a unified decision-making framework.

The primary goal of this component is to identify the best lane-change and velocity actions for each state by combining the symbolic rules with background knowledge. This component prioritizes rules based on their criticality: (1) \textit{fatal} lane changing rules (\texttt{h1-h2}) identify deadly actions, which are eliminated from the action space. (2) \textit{Risky} lane changing rules (\texttt{h3-h5}) detect risky actions; these actions are removed from the action space but retained as backup options if no other actions remain. (3) Finally, in the refined action space, \textit{efficiency} (\texttt{h6-h7}) rules select the action leading to the most efficient path for the AV. Additionally, \textit{smoothness} (\texttt{h8-h10}) rules adjust the longitudinal velocity to create a smooth driving.

Supplementary rules are also introduced to ensure consistent reasoning across the hypotheses, enabling seamless interpretation and coordination of the high-level symbolic outputs. These outputs are then used to control the AV via a low-level controller.

To determine the best action for a given state $S_t$, the framework first applies rules \verb|h1| through \verb|h5| to eliminate actions deemed fatal or risky, thus narrowing the action space to only safe candidates. From the remaining options, the most efficient action is selected using rules \verb|h6| and \verb|h7|, which assess the relative quality and priority of the remaining lane-change options. When multiple candidate rules are triggered simultaneously, the system applies a deterministic priority order based on rule specificity. In practice, this means that rules supported by larger numbers of positive examples take precedence. For instance, left-lane change rules are prioritized over right-lane change rules. The final decision yields the optimal action $a_t$, which can be LK, LLC, or RLC. As illustrated in Fig.~\ref{sil}, once $a_t$ is chosen, the AV's lane ID is updated via a switch box mechanism that maps the decision to a corresponding lateral position $y_d^{NL}$. A low-level controller is then used to generate the lateral acceleration command $a_y$, where a proportional-integral-derivative (PID) controller is employed as in~\cite{sharifi2025toward}. The controller gains $K_p^y$, $K_I^y$, and $K_D^y$ are 4, 1, and 0, respectively.

For longitudinal velocity control, the current state $S_t$ is evaluated against rules \verb|h8| to \verb|h10| to determine which of the three acceleration phases—\verb|catch-up|, \verb|follow-up|, or \verb|brake|—is applicable. Based on the selected rule and the formulation in Eq.~\ref{accelerations}, the required longitudinal acceleration $a_x$ is computed. The desired velocity is then calculated as $v^d_x = v_x + a_x \Delta t$ and passed as input to a PID controller responsible for regulating thrust. For this task, the controller gains $K_p^x$, $K_I^x$, and $K_D^x$ are 3, 1, and 0, respectively. The PID controller ensures smooth and continuous velocity control in accordance with the selected symbolic rule.

In summary, the SIL framework integrates symbolic reasoning and control by mapping learned high-level hypotheses into executable actions. These actions are aligned with the safety, efficiency, and smoothness principles established throughout the rule induction process.

%%%%%%%%%%%%%%%%%%%%%%%%%%%%%%%%%%%%%%%%%%
\section{Results}
\label{sec:results}

This section presents the performance evaluation of the proposed method in comparison with several imitation learning baselines. The experiments are conducted using the HighD~\cite{krajewski2018highd} and NGSim~\cite{yeo2008oversaturated} datasets, which offer realistic driving scenarios to evaluate both learning and generalization capabilities.

\subsection{\textbf{Baselines}}

\subsubsection{DNNIL: Deep-Neural-Network-based Imitation Learning}

We implemented a fully connected deep neural network (DNN) as the baseline for DNNIL. The network architecture consisted of three layers: the first two layers contained 128 nodes each, while the final layer had three output nodes representing the action space. A softmax activation function was applied in the output layer. The network was implemented using the PyTorch library, and the ADAM optimization algorithm was used with a learning rate of $\alpha = 1\text{e}^{-4}$ and a batch size of 256.

To train the network, we collected over 160{,}000 state-action pairs from the HighD dataset by tracking all vehicles from the beginning to the end of each sequence. Each state was represented by the normalized relative positions of eight surrounding target vehicles (TVs) and the normalized velocity of the AV, with normalization performed using the predefined maximum values in the HighD dataset (118.8~km/h for velocity and 250~m for distance). Lane-change actions were detected by comparing the initial and final lane indices of each vehicle within each frame. Based on this detection, we encoded the action as a binary list of length three, indicating the corresponding maneuver.

It is important to note that this setup focused exclusively on imitating lane-change decisions. Velocity control was handled separately using rule-based methods rather than being learned by the network. This design aligns with the core objective of comparing symbolic policy learning with black-box neural imitation learning in the context of high-level decision-making.

In addition to DNNIL, we implemented two other well-established imitation learning methods on the HighD dataset to further validate the reliability and effectiveness of the proposed approach. These baselines are briefly described below.

\subsubsection{BCMDN: Behavioral Cloning with Mixture Density Networks}

This method combines Behavior Cloning (BC)~\cite{morga2024behavioral}—a widely used approach in which an agent learns to mimic expert behavior—with Mixture Density Networks (MDNs)~\cite{bishop1994mixture}, a neural architecture designed to model probability distributions. Unlike traditional BC, which predicts a single deterministic action, BCMDN predicts a probability distribution over multiple possible actions using MDNs. This modeling of uncertainty allows the agent to represent multiple plausible behaviors in ambiguous or highly variable situations, thereby improving robustness and flexibility in complex driving scenarios.

\subsubsection{GAIL: Generative Adversarial Imitation Learning}

Generative Adversarial Imitation Learning (GAIL)~\cite{ho2016generative} is an advanced imitation learning framework built on Generative Adversarial Networks (GANs)~\cite{goodfellow2020generative}. In this approach, a discriminator network is trained to distinguish between expert actions and those produced by a learning agent. Simultaneously, a generator (policy) network attempts to generate actions that are indistinguishable from expert actions according to the discriminator. GAIL optimizes the policy such that it closely mimics expert behavior, all without requiring explicitly defined reward signals~\cite{ho2016generative}.

For further details regarding the implementation of BCMDN and GAIL in the context of driving behavior modeling on the HighD dataset, we refer interested readers to~\cite{kuutti2021adversarial, yildirim2023human}.

\subsection{\textbf{Symbolic Imitation Learning Implementation}}

In contrast to the DNNIL method, the SIL framework relies on a relatively small number of positive and negative examples to learn unknown rules, as summarized in Table~\ref{rules}. Once constructed, the SIL framework is capable of making interpretable decisions in each state by leveraging symbolic first-order logic. These decisions are also generalizable, as the learned rules are not constrained to specific training scenarios.

As shown in Table~\ref{rules}, another advantage of the SIL method is its computational efficiency—each rule can be induced in a fraction of a second (see $T_i$ column). For implementation, we used the HighD dataset to simulate a virtual AV agent initialized with arbitrary lane and velocity values. The agent was then tasked with driving safely, efficiently, and smoothly on the highway using the learned symbolic policies.

%Similar to DNNIL, the SIL training process is conducted offline and completed prior to deployment. However, unlike neural networks that often require large datasets and extended training time, the ILP-based SIL framework is highly sample-efficient and can often learn meaningful rules from only a few well-labeled examples.

Similar to DNNIL, the SIL training process is conducted offline and completed prior to deployment. However, unlike neural networks that often require large datasets and extended training time, the ILP-based SIL framework is highly sample-efficient and can often learn meaningful rules from only a few well-labeled examples. In contrast, the baselines required substantially longer training times: approximately 13 hours for DNNIL, 14 hours for BCMDN, and 20 hours for GAIL on a workstation equipped with an RTX-series GPU, 32\,GB RAM, and an Intel i7 processor.

\begin{table}[H]
\caption{Comparison of SIL and baseline methods on evaluation episodes from the HighD and NGSim scenarios with different seeds.}

\label{tabresults}
\begin{adjustwidth}{-\extralength}{0cm}
\begin{tabularx}{\fulllength}{lCCCCCCCCc}
\toprule
\textbf{Direction} & \multicolumn{4}{c}{\textbf{Left-to-Right (HighD)}} & \multicolumn{4}{c}{\textbf{Right-to-Left (HighD)}} & \textbf{NGSim} \\
\cmidrule(l){1-1} \cmidrule(lr){2-5} \cmidrule(lr){6-9} \cmidrule(l){10-10}
\textbf{Method} & \textbf{DNNIL} & \textbf{BCMDN} & \textbf{GAIL} & \textbf{SIL} & \textbf{DNNIL} & \textbf{BCMDN} & \textbf{GAIL} & \textbf{SIL} & \textbf{SIL} \\
\midrule
$\textbf{N}_{\textbf{LC}}$        & 3  & 2  & 31    & 40    & 4  & 5  & 23    & 38    & 34 \\
$\bar{\textbf{T}}\,\textbf{(s)}$    & 65.28  & 69.67  & 56.78  & 64.84  & 65.13  & 76.5   & 50.76  & 64.93  & 76.38 \\
$\bar{\textbf{V}}\,\textbf{(km/h)}$ & 110.18 & 104.29    & 117.65    & 116.59 & 109.22 & 95.5     & 112.5     & 115.72  & 98.00 \\
$\textbf{SR}_{\textbf{d}}\textbf{ (\%)}$    & 95.1  & 96  & 88.3  & \textbf{100}  & 94  & 96.6  & 75.6  & \textbf{99.3}  & \textbf{99} \\
$\textbf{SR}_{\textbf{c}}\textbf{ (\%)}$      & 88  & 88  & 46    & \textbf{100}     & 84  & 86  & 36    & \textbf{98}     & \textbf{96} \\
\bottomrule
\end{tabularx}
\end{adjustwidth}
\noindent{\footnotesize $N_{LC}$: Number of lane changes; $\text{SR}_c$: Collision-based success rate; $\text{SR}_d$: Distance-based success rate; $\bar{T}$: Average mission time; $\bar{V}$: Average agent speed.}
\end{table}

\section{Discussion}
\label{discussion}

All methods were evaluated under comparable conditions to facilitate a fair comparison of their overall performance. Each agent was initialized with similar positions, velocities, and driving directions. To assess the effectiveness of each approach, we defined performance metrics in three key categories: safety, efficiency, and smoothness. Safety was evaluated by the two success rates: collision-based success rate $\text{SR}_c$ and distance-based success rate $\text{SR}_d$, which are computed as:
\begin{equation}
    \text{SR}_c = (1-\frac{N_C}{N}) \times 100, \quad \text{SR}_d = \frac{\bar{D}}{L}\times100,
\end{equation}
where $N_C$ is the number of collisions, and $N$ is the number of evaluation episodes. Also, $\bar{D}$ is the average traveled distance and $L$ is the total length of the driving scenario. $\text{SR}_c$ indicates the percentage of the episodes completed without a collision while $\text{SR}_d$ indicates the percentage of the traveled distance compared to the total traveling distance. Higher values of both metrics show a higher degree of safety.

Furthermore, the efficiency and smoothness were evaluated through the number of lane changes ($N_{LC}$). Additionally, average agent speed ($\bar{V}$) was used as a composite measure of efficiency, calculated as $\bar{V} = \frac{\bar{D}}{\bar{T}}$, where $\bar{T}$ denotes mission time per episode.

Using the HighD dataset, we tested all methods over $N=100$ episodes with at least five different seeds, with each episode consisting of a $L = 2,100 \,(m)$ driving track. For each experiment, the AV operated in either the left-to-right (L2R) or right-to-left (R2L) direction. All agents were trained exclusively in the L2R direction, and their generalization was assessed by evaluating performance in the R2L direction and the NGSim environment. The average comparative results over seeds are summarized in Table~\ref{tabresults}.

\textbf{Safety Analysis:} As shown in Table~\ref{tabresults}, the SIL agent completed all evaluation scenarios with 100$\%$ collision-based success rate in the L2R scenario, demonstrating strong safety performance. In contrast, the corresponding success rate for the DNNIL, BCMDN, and GAIL are significantly smaller. Due to the explicit safety rules considered in the proposed framework, the SIL agent complies with the safety rules and avoid fatal and risky lane changes which lead to collisions. Moreover, in the R2L scenario and the NGSim scenarios, it consistently maintains superiority over the baselines by success rates 98$\%$ and 96$\%$, respectively.

From the distance-based success rate perspective, the SIL agent similarly outperforms the baselines by completing $100\%$ and $99.3\%$ of the path on average in L2R and R2L directions, respectively. These superior results stems not only from explicit safety rules but also from the longitudinal velocity rules, which ensure safe distances from the front and back intruder vehicles.

% this agent experienced 8 collisions out of 100 episodes in the unseen R2L direction. BCMDN showed similar limitations, with 7 collisions out of 100 episodes in R2L, suggesting poor adaptability to unfamiliar traffic conditions. While the GAIL agent achieved a higher number of lane changes—particularly in the L2R direction—it also suffered from a high collision rate, with 32 collisions out of 100 episodes in R2L, highlighting its difficulty in balancing proactive maneuvers with safety constraints. In comparison, the SIL agent maintained only 1 collision out of 100 episodes in R2L, illustrating superior generalizability.

\textbf{Efficiency Analysis:} 
To assess the efficiency of the agents, we examined their average speed per episode, denoted by $\bar{V}$. This metric serves as a proxy for effective lane-change behavior, as faster travel generally correlates with successful overtaking. We computed $\bar{V}$ across both directions for all agents and used the average value for comparison. The results show that DNNIL, BCMDN, GAIL, and SIL achieved $\bar{V}$ values of 109.7, 99.89, 115.07, and 116.06 km/h, respectively. Notably, the SIL agent achieved the highest average speed, suggesting that its rule-based lane changes allowed it to find free lanes efficiently and maintain higher velocities, ultimately reducing overall travel time.

\textbf{Smoothness Analysis:}
To analyze the smoothness, we consider the the number of lane changes ($N_{LC}$), which was close to zero for the DNNIL and BCMDN agents, indicating their limited ability to learn and execute lane-change maneuvers effectively. While these agents exhibited smooth driving behavior, they often remained in a single lane throughout the episode. This tendency not only reduces responsiveness but may also contribute to traffic congestion due to inefficient lane utilization. Despite being trained on a large volume of data, the inability of these models to generalize lane-change behaviors highlights their inherent sample inefficiency.

The GAIL agent, in contrast, performed a significantly higher number of lane changes. However, this came at the cost of a high collision rate, indicating that its behavior, while active, was not reliably safe. On the other hand, the SIL agent changed lanes approximately once per episode, guided by explicitly learned efficient lane changing rules. These rules enabled the AV to make strategic, context-aware lane changes, which contributed to shorter travel times without compromising safety.

%To evaluate the generalizability of the SIL framework, we evaluationed it on a different dataset—NGSim~\cite{yeo2008oversaturated}—which captures diverse urban highway scenarios in the United States. Among the available scenarios, the US-101 subset was selected for evaluation due to its highway characteristics, which are broadly similar to the HighD dataset but vary in vehicle density, lane structure, and speed distribution. Because the NGSim environment is highly congested, we reduced the number of vehicles to create feasible driving space for the AV. The state space representation was unified between the HighD and NGSim datasets.The reduction in the number of vehicles in NGSim was performed to align with the same eight-sector representation used in HighD. As shown in Table~\ref{tabresults}, the SIL agent continued to perform safely, maintaining a relatively low number of collisions despite the shift in environment and dynamics.

\textbf{Generalizability Analysis:} To assess the generalizability of the SIL framework, we tested it on a different dataset—NGSim~\cite{yeo2008oversaturated}—which captures diverse urban highway scenarios in the United States. Among the available subsets, US-101 was chosen because its highway characteristics are broadly comparable to those of the HighD dataset, while differing in vehicle density, lane structure, and speed distribution. Given that the NGSim environment is highly congested, we reduced the number of vehicles to ensure feasible driving space for the AV while preserving comparability. To this end, the state-space representation was kept consistent across both datasets by maintaining the same eight-sector structure as in HighD. As shown in Table~\ref{tabresults}, the SIL agent continued to perform safely, maintaining a low number of collisions despite the changes in environment and traffic dynamics.

Notably, some collisions observed in the NGSim experiments were unavoidable. This is primarily due to the fact that surrounding vehicles in the dataset are not aware of the SIL-controlled AV and thus do not respond to its presence. As a result, collisions often occurred from the rear, where other vehicles failed to maintain a safe following distance. Additionally, the AV's average velocity in the NGSim scenarios was lower compared to the HighD due to the overall slower traffic flow and higher congestion levels in the NGSim dataset. Nevertheless, the ILP-generated rules enabled the AV to adapt effectively, demonstrating that the SIL framework is robust to variations in lane configurations and speed distributions.

Another advantage of the SIL framework lies in its computational efficiency. As shown in Table~\ref{rules}, each rule can be induced in a relatively short time, whereas training a DNNIL model on large datasets demands substantial time and computational resources. Moreover, in many real-world applications, large volumes of labeled data may not be readily available. In contrast, the SIL framework can learn effectively from a small set of human-labeled examples, making it more practical and accessible. Furthermore, the explicit, interpretable nature of the SIL rule base aligns well with established automotive functional safety standards, such as ISO 26262 \cite{palin2011iso}, by addressing both functional safety and AI system assurance requirements.

Despite these strengths, the proposed method has several limitations. One notable drawback is that all datasets should be carefully labeled to avoid noise issues associated with classical ILP systems. The presence of incorrectly labeled positive or negative examples can significantly hinder the rule induction process. Additionally, as discussed in subsection~\ref{data-acq}, each unknown rule must be learned under a carefully defined setting, including a tailored bias set and example set, which adds to the system design complexity. Another challenge arises when attempting to extract rules from real-world human driving data. Human drivers often take different actions in similar situations due to personal preferences or unobservable knowledge, making it difficult to infer consistent rules. 

Future work should aim to address these challenges by improving the noise tolerance of ILP systems and exploring semi-automated ways to configure rule learning environments. Advancing in these directions would help scale the application of symbolic imitation learning to more diverse and unstructured real-world driving scenarios.

%%%%%%%%%%%%%%%%%%%%%%%%%%%%%%%%%%%%%%%%%%
\section{Conclusion} \label{conclusion}

This paper introduced a novel Symbolic Imitation Learning (SIL) framework that utilizes human driving background knowledge and example-based reasoning to extract interpretable decision-making rules for autonomous vehicles via Inductive Logic Programming (ILP). The proposed method is structured around three core components: knowledge acquisition, rule induction, and rule aggregation. These components work in tandem to enable the derivation and integration of symbolic rules that guide autonomous driving behavior. Through extensive experiments on the HighD and NGSim datasets, we demonstrated that SIL outperforms prominent deep neural network-based imitation learning approaches in key performance dimensions, including safety, efficiency, and smoothness. Importantly, the method offers full interpretability by operating within a first-order logic framework and exhibits strong generalizability in previously unseen environments. In addition, SIL is highly sample-efficient, requiring substantially fewer state-action pairs to learn effective policies compared to black-box learning methods.

Future work should explore the application of the SIL framework in more complex and dynamic driving contexts, such as highway merging, bidirectional traffic, and scenarios involving heterogeneous driving behaviors and intentions. While the extracted rules have proven effective in highway-like environments, extending the rule base to accommodate broader traffic situations remains a valuable direction for research and development.

%%%%%%%%%%%%%%%%%%%%%%%%%%%%%%%%%%%%%%%%%%
%\section{Patents}

%This section is not mandatory, but may be added if there are patents resulting from the work reported in this manuscript.

%%%%%%%%%%%%%%%%%%%%%%%%%%%%%%%%%%%%%%%%%%
\vspace{6pt} 

%%%%%%%%%%%%%%%%%%%%%%%%%%%%%%%%%%%%%%%%%%
%% optional
%\supplementary{The following supporting information can be downloaded at:  \linksupplementary{s1}, Figure S1: title; Table S1: title; Video S1: title.}


% Only used for preprtints:
% \supplementary{The following supporting information can be downloaded at the website of this paper posted on \href{https://www.preprints.org/}{Preprints.org}.}

%\begin{tabularx}{\textwidth}{lll}
%\toprule
%\textbf{Name} & \textbf{Type} & \textbf{Description} \\
%\midrule
%S1 & Python script (.py) & Script of python source code used in XX \\
%S2 & Text (.txt) & Script of modelling code used to make Figure X \\
%S3 & Text (.txt) & Raw data from experiment X \\
%S4 & Video (.mp4) & Video demonstrating the hardware in use \\
%\bottomrule
%\end{tabularx}
%}

%%%%%%%%%%%%%%%%%%%%%%%%%%%%%%%%%%%%%%%%%%
\authorcontributions{%For research articles with several authors, a short paragraph specifying their individual contributions must be provided. The following statements should be used ``
Conceptualization, I.S. and M.Y.; methodology, I.S.; software, I.S.; validation, I.S. and M.Y.; data curation, I.S. and M.Y; writing---original draft preparation, I.S.; writing---review and editing, I.S. and M.Y; visualization M.Y and I.S.; supervision, S.F.; project administration, S.F. . All authors have read and agreed to the published version of the manuscript.
%'', please turn to the  \href{http://img.mdpi.org/data/contributor-role-instruction.pdf}{CRediT taxonomy} for the term explanation. Authorship must be limited to those who have contributed substantially to the work~reported.
}% formal analysis, X.X.; investigation, X.X.; resources, X.X.;

\funding{%Please add: ``
This research received no external funding.
%'' or ``This research was funded by NAME OF FUNDER grant number XXX.'' and  and ``The APC was funded by XXX''. Check carefully that the details given are accurate and use the standard spelling of funding agency names at \url{https://search.crossref.org/funding}, any errors may affect your future funding.
}

\institutionalreview{%In this section, you should add the Institutional Review Board Statement and approval number, if relevant to your study. You might choose to exclude this statement if the study did not require ethical approval. Please note that the Editorial Office might ask you for further information. Please add “The study was conducted in accordance with the Declaration of Helsinki, and approved by the Institutional Review Board (or Ethics Committee) of NAME OF INSTITUTE (protocol code XXX and date of approval).” for studies involving humans. OR “The animal study protocol was approved by the Institutional Review Board (or Ethics Committee) of NAME OF INSTITUTE (protocol code XXX and date of approval).” for studies involving animals. OR “Ethical review and approval were waived for this study due to REASON (please provide a detailed justification).” OR “
Not applicable.
%” for studies not involving humans or animals.
}

\informedconsent{%Any research article describing a study involving humans should contain this statement. Please add ``Informed consent was obtained from all subjects involved in the study.'' OR ``Patient consent was waived due to REASON (please provide a detailed justification).'' OR ``
Not applicable.
%'' for studies not involving humans. You might also choose to exclude this statement if the study did not involve humans.

%Written informed consent for publication must be obtained from participating patients who can be identified (including by the patients themselves). Please state ``Written informed consent has been obtained from the patient(s) to publish this paper'' if applicable.
}

%\dataavailability{We encourage all authors of articles published in MDPI journals to share their research data. In this section, please provide details regarding where data supporting reported results can be found, including links to publicly archived datasets analyzed or generated during the study. Where no new data were created, or where data is unavailable due to privacy or ethical restrictions, a statement is still required. Suggested Data Availability Statements are available in section ``MDPI Research Data Policies'' at \url{https://www.mdpi.com/ethics}.} 

\dataavailability{The implementation of the proposed methods is available at: \href{https://github.com/CAV-Research-Lab/Symbolic-Imitation-Learning}{github.com/CAV-Research-Lab/Symbolic-Imitation-Learning}.\\  
The HighD dataset can be accessed at: \href{https://www.highd-dataset.com}{highd-dataset.com}, and the NGSim dataset is available at: \href{https://ops.fhwa.dot.gov/trafficanalysistools/ngsim.htm}{ops.fhwa.dot.gov/trafficanalysistools}.}

\acknowledgments{ During the preparation of this manuscript, the authors partially used GPT-4o for language editing and proofreading. The authors have reviewed and edited the output and take full responsibility for the content of this publication.}

\conflictsofinterest{%Declare conflicts of interest or state ``
The authors declare no conflicts of interest.
%'' Authors must identify and declare any personal circumstances or interest that may be perceived as inappropriately influencing the representation or interpretation of reported research results. Any role of the funders in the design of the study; in the collection, analyses or interpretation of data; in the writing of the manuscript; or in the decision to publish the results must be declared in this section. If there is no role, please state ``The funders had no role in the design of the study; in the collection, analyses, or interpretation of data; in the writing of the manuscript; or in the decision to publish the results''.
} 

%%%%%%%%%%%%%%%%%%%%%%%%%%%%%%%%%%%%%%%%%%
%% Optional

%% Only for journal Encyclopedia
%\entrylink{The Link to this entry published on the encyclopedia platform.}

\abbreviations{Abbreviations}{
The following abbreviations are used in this manuscript:

\noindent 
\begin{tabular}{@{}ll}
AD    & Autonomous Driving\\
AV    & Autonomous Vehicle\\
DNNIL   & Deep-Neural-Network-based Imitation Learning\\
IL    & Imitation Learning\\
ILP   & Inductive Logic Programming\\
FOL   & First-Order Logic\\
HighD & HighD Traffic Dataset\\
L2R   & Left to Right\\
LK    & Lane Keeping\\
LLC   & Left Lane Change\\
NGSIM & Next Generation Simulation Dataset\\
PID   & Proportional-Integral-Derivative\\
R2L   & Right to Left\\
RLC   & Right Lane Change\\
SIL   & Symbolic Imitation Learning\\
TV    & Target Vehicle\\
XAI   & Explainable Artificial Intelligence\\
\end{tabular}
}

%%%%%%%%%%%%%%%%%%%%%%%%%%%%%%%%%%%%%%%%%%
%% Optional
\begin{comment}
    
\appendixtitles{no} % Leave argument "no" if all appendix headings stay EMPTY (then no dot is printed after "Appendix A"). If the appendix sections contain a heading then change the argument to "yes".
\appendixstart
\appendix
\section[\appendixname~\thesection]{}
\subsection[\appendixname~\thesubsection]{}
The appendix is an optional section that can contain details and data supplemental to the main text---for example, explanations of experimental details that would disrupt the flow of the main text but nonetheless remain crucial to understanding and reproducing the research shown; figures of replicates for experiments of which representative data are shown in the main text can be added here if brief, or as Supplementary Data. Mathematical proofs of results not central to the paper can be added as an appendix.

\begin{table}[H] 
\caption{This is a table caption.\label{tab5}}
%\newcolumntype{C}{>{\centering\arraybackslash}X}
\begin{tabularx}{\textwidth}{CCC}
\toprule
\textbf{Title 1}	& \textbf{Title 2}	& \textbf{Title 3}\\
\midrule
Entry 1		& Data			& Data\\
Entry 2		& Data			& Data\\
\bottomrule
\end{tabularx}
\end{table}

\section[\appendixname~\thesection]{}
All appendix sections must be cited in the main text. In the appendices, Figures, Tables, etc. should be labeled, starting with ``A''---e.g., Figure A1, Figure A2, etc.
\end{comment}

%%%%%%%%%%%%%%%%%%%%%%%%%%%%%%%%%%%%%%%%%%
%\isPreprints{}{% This command is only used for ``preprints''.
\begin{adjustwidth}{-\extralength}{0cm}

\reftitle{References}
\bibliography{references}
\PublishersNote{}

\end{adjustwidth}

%\isPreprints{}{% This command is only used for ``preprints''.
\appendix
\section*{Appendix}
\subsection{Symbolic Predicate Definitions}
\begin{table}[H]
\centering
\renewcommand{\arraystretch}{1.2}
% \begin{adjustwidth}{-\extralength}{0cm}
\caption{List of symbolic predicates used in the SIL framework and their definitions.}
\label{tab:predicate-definitions}
\begin{tabularx}{\textwidth}{lX}
\toprule
\textbf{Predicate} & \textbf{Definition} \\
\midrule
\verb|front_isBusy| & Indicates whether the front sector is occupied by a target vehicle. \\
\verb|frontRight_isBusy| & Indicates whether the front-right sector is occupied. \\
\verb|right_isBusy| & Indicates whether the right sector is occupied. \\
\verb|backRight_isBusy| & Indicates whether the back-right sector is occupied. \\
\verb|back_isBusy| & Indicates whether the back sector is occupied. \\
\verb|backLeft_isBusy| & Indicates whether the back-left sector is occupied. \\
\verb|left_isBusy| & Indicates whether the left sector is occupied. \\
\verb|frontLeft_isBusy| & Indicates whether the front-left sector is occupied. \\
\verb|right_isValid| & Indicates whether the right lane is valid (within road boundaries). \\
\verb|left_isValid| & Indicates whether the left lane is valid. \\
\verb|frontVel_isBigger| & The front vehicle is moving faster than the AV. If $|v_{x,TV}-v_{x,AV}| > \eta_v$ and $v_{x,TV} > v_{x,AV}$, it is true; otherwise, it is false. \\
\verb|frontVel_isEqual| & The front vehicle is moving at a similar speed as the AV. If $|v_{x,TV}-v_{x,AV}| < \eta_v$, it is true; otherwise, it is false.  \\
\verb|frontVel_isLower| & The front vehicle is moving slower than the AV. If $|v_{x,TV}-v_{x,AV}| > \eta_v$ and $v_{x,TV} < v_{x,AV}$, it is true; otherwise, it is false. \\
\verb|frontRightVel_isBigger| & The vehicle in the front-right is faster than the AV. \\
\verb|frontRightVel_isLower| & The vehicle in the front-right is slower than the AV. \\
\verb|backRightVel_isBigger| & The vehicle in the back-right is faster than the AV. \\
\verb|backLeftVel_isBigger| & The vehicle in the back-left is faster than the AV. \\
\verb|frontLeftVel_isLower| & The vehicle in the front-left is slower than the AV. \\
\verb|backVel_isBigger| & The vehicle in the back sector is faster than the AV. \\
\verb|backDist_isSafe| & The distance between the AV and the vehicle in the back is safe. If the distance is more than $C$, it is true; otherwise, it is false. \\
\verb|frontDist_isSafe| & The distance to the front vehicle is within a safe threshold. If the distance is more than $C$, it is true; otherwise, it is false. \\
\verb|rlc_isFatal| & Indicates the right lane change (RLC) is fatal. \\
\verb|llc_isFatal| & Indicates the left lane change (LLC) is fatal. \\
\verb|lk_isRisky| & Indicates lane keeping (LK) is risky in the current state. \\
\verb|rlc_isRisky| & Indicates RLC is risky in the current state. \\
\verb|llc_isRisky| & Indicates LLC is risky in the current state. \\
\verb|llc_isBetter| & Indicates LLC is more efficient than RLC in the given state. \\
\verb|rlc_isBetter| & Indicates RLC is more efficient than LLC. \\
\verb|reachDesiredSpeed| & Indicates the AV should accelerate to reach its desired speed. \\
\verb|reachFrontSpeed| & Indicates the AV should adjust speed to match the front vehicle. \\
\verb|brake| & Indicates that the AV should decelerate due to an unsafe distance ahead. \\
\bottomrule
\end{tabularx}
% \end{adjustwidth}
\end{table}

\subsection{Experimental Results with Noisy Datasets}
\begin{table}[!h]
\centering
\caption{Experimental results with noisy datasets. All noises applied with seed 42. In Popper, parameter $b_{\text{max}}$ indicates the maximum number of body predicates for each rule, which directly affects the rule space size.}
\label{tab:noisy_results}
\renewcommand{\arraystretch}{1.2}
\begin{tabular}{ccccccc}
\toprule
\makecell{\textbf{Dataset} \\ \textbf{Category}} & \textbf{Action} & \textbf{Noise (\%)} & \textbf{$\textbf{b}_{\text{max}}$} & \textbf{Precision} & \textbf{Recall} & \textbf{$\textbf{T}_i$ (sec)} \\
\midrule
\multirow{8}{*}{\rotatebox{90}{\textbf{Fatal Lane Changing}}} &  \multirow{4}{*}{\textbf{RLC}}   
                 & 2  & 2 & 0.98 & 0.99 & 0.24 \\
           &     & 5  & 2 & 0.95 & 0.98 & 0.25 \\
           &     & 10 & 2 & 0.90 & 0.97 & 0.27 \\
           &     & 20 & 2 & 0.79 & 0.94 & 0.36 \\

\cline{2-7}
           &  \multirow{4}{*}{\textbf{LLC}}   
                 & 2 & 2 & 0.98 & 0.99 & 0.24 \\
           &     & 5  & 2 & 0.95    & 0.98 & 0.26 \\
           &     & 10 & 2 & 0.91    & 0.95 & 0.28 \\
           &     & 20 & 2 & 0.82    & 0.91 & 0.36 \\
\midrule

\multirow{12}{*}{\rotatebox[origin=c]{90}{\textbf{Risky Lane Changing}}} &  \multirow{4}{*}{\textbf{RLC}}   

                 & 2  & 3 & 0.99 & 0.82 & 1.52 \\
           &     & 5  & 3 & 0.97 & 0.64 & 1.83 \\
           &     & 10 & 3 & 0.90 & 0.46 & 1.90 \\
           &     & 20 & 3 & 0.78 & 0.27 & 2.20 \\

\cline{2-7}
           &  \multirow{4}{*}{\textbf{LLC}}   
                 & 2  & 3 & 0.98 & 0.96 & 1.88 \\
           &     & 5  & 3 & 0.95 & 0.90 & 2.31 \\
           &     & 10 & 3 & 0.92 & 0.80 & 2.92 \\
           &     & 20 & 3 & 0.82 & 0.64 & 0.49 \\
\cline{2-7}
           &  \multirow{4}{*}{\textbf{LLC}}   
                 & 2  & 3 & 0.98 & 0.95 & 1.82 \\
           &     & 5  & 3 & 0.94 & 0.90 & 2.82 \\
           &     & 10 & 3 & 0.91 & 0.79 & 2.71 \\
           &     & 20 & 2 & 0.82 & 0.63 & 0.53 \\
\midrule
\multirow{8}{*}{\rotatebox[origin=c]{90}{\textbf{Efficient Lane Changing}}} &  \multirow{4}{*}{\textbf{RLC}}   
                 & 2  & 5 & 0.97 & 0.62 & 0.82 \\
           &     & 5  & 5 & 0.97 & 0.38 & 1.41 \\
           &     & 10 & 5 & 0.91 & 0.23 & 2.80 \\
           &     & 20 & 5 & 0.88 & 0.12 & 3.80 \\

\cline{2-7}
           &  \multirow{4}{*}{\textbf{LLC}}   
                 & 2  & 5 & 0.98 & 0.88 & 1.91 \\
           &     & 5  & 5 & 0.97 & 0.73 & 2.67 \\
           &     & 10 & 5 & 0.88 & 0.56 & 3.58 \\
           &     & 20 & 5 & 0.80 & 0.36 & 4.72 \\
\midrule
\multirow{12}{*}{\rotatebox[origin=c]{90}{\textbf{Smooth Longitudinal Velocity}}} &  \multirow{4}{*}{\textbf{Catch-up}}   
                 & 2  & 3 & 0.98 & 0.98 & 2.80 \\
           &     & 5  & 2 & 0.94 & 0.96 & 0.48 \\
           &     & 10 & 2 & 0.91 & 0.89 & 1.62 \\
           &     & 20 & 1 & 0.79 & 0.80 & 0.26 \\

\cline{2-7}
           &  \multirow{4}{*}{\textbf{Follow-up}}   
                 & 2  & 4 & 0.98 & 0.94 & 18.84 \\
           &     & 5  & 4 & 0.95 & 0.86 & 21.87 \\
           &     & 10 & 4 & 0.92 & 0.74 & 40.08 \\
           &     & 20 & 3 & 0.83 & 0.57 & 79.55 \\
\cline{2-7}
           &  \multirow{4}{*}{\textbf{Brake}}   
                 & 2  & 3 & 0.98 & 0.95 & 2.07 \\
           &     & 5  & 3 & 0.95 & 0.86 & 2.72 \\
           &     & 10 & 3 & 0.91 & 0.75 & 38.54 \\
           &     & 20 & 2 & 0.84 & 0.56 & 0.92 \\

\bottomrule
\end{tabular}
\end{table}

\subsection{BCMDN and GAIL Pseudo Codes}

\begin{algorithm}[h]
    \caption{Behavior Cloning with a Mixture Density Network (BCMDN)}
    \label{algo:bc_mdn}
    \begin{algorithmic}[1]
        \REQUIRE Dataset $\mathcal{D}=\{(s_i,a_i)\}_{i=1}^N$ of expert state--action pairs
        \REQUIRE Number of Gaussian components $K$, learning rate $\alpha$, batch size $B$
        \ENSURE Policy $\pi_\Theta(a\,|\,s)$ parameterized by MDN $\Theta=\{\phi,\psi\}$

        \STATE \textbf{Model:} Feature encoder $h_\phi(s)$ (e.g., MLP/CNN); MDN head $g_\psi(\cdot)$ outputs per state $s$:
        \STATE \hspace{1em}$\{\boldsymbol{\pi}(s), \boldsymbol{\mu}(s), \boldsymbol{\sigma}(s)\} = g_\psi(h_\phi(s))$
        \STATE \hspace{1em}\textit{Constraints:} $\pi_k=\mathrm{softmax}(z^\pi_k/\tau)$, $\sigma_{k,d}=\mathrm{softplus}(z^\sigma_{k,d})+\epsilon$
        \STATE Initialize parameters $\Theta\leftarrow\Theta_0$
        \STATE Normalize/standardize states and actions (fit on $\mathcal{D}$)

        \FOR{\textbf{epoch} $=1,2,\ldots$}
            \STATE Shuffle $\mathcal{D}$ and split into mini-batches $\{\mathcal{B}_j\}$
            \FOR{each mini-batch $\mathcal{B}=\{(s,a)\}_{b=1}^B$}
                \STATE Compute MDN outputs $\{\pi_k,\mu_{k},\sigma_{k}\}_{k=1}^K$ for each $(s,a)\in\mathcal{B}$
                \STATE \textbf{Negative log-likelihood (per sample):}
                \STATE \hspace{1em}$\displaystyle
                    \mathcal{L}_{\text{NLL}}(s,a)=
                    -\log \Bigg( \sum_{k=1}^{K} \pi_k(s)\;
                        \prod_{d=1}^{D} \mathcal{N}\big(a_d \,\big|\, \mu_{k,d}(s),\, \sigma_{k,d}^2(s)\big)
                    \Bigg)$
                \STATE \textbf{Regularization (optional):}
                \STATE \hspace{1em}$\mathcal{L}_{\text{reg}}=\beta \|\Theta\|_2^2 \;-\; \eta\, H(\boldsymbol{\pi}(s))$ \hfill 
                \STATE \textbf{Batch loss:} $\displaystyle \mathcal{L}=\frac{1}{B}\sum_{(s,a)\in\mathcal{B}}\mathcal{L}_{\text{NLL}}(s,a)+\mathcal{L}_{\text{reg}}$
                \STATE Update $\Theta \leftarrow \Theta - \alpha\, \nabla_\Theta \mathcal{L}$
            \ENDFOR
        \ENDFOR
        \STATE \textbf{Inference (deterministic):} $\displaystyle a^\star(s)=\sum_{k=1}^{K}\pi_k(s)\,\mu_k(s)$
        \STATE \textbf{Inference (stochastic):} sample $k\sim \mathrm{Cat}(\boldsymbol{\pi}(s))$, then $a\sim \mathcal{N}(\mu_k(s),\mathrm{diag}(\sigma_k^2(s)))$
        \STATE \textbf{return} MDN policy $\pi_\Theta(a\,|\,s)$
    \end{algorithmic}
\end{algorithm}

\begin{algorithm}[h]
    \caption{Generative Adversarial Imitation Learning (GAIL) with PPO}
    \label{algo:gail}
    \begin{algorithmic}[1]
        \REQUIRE Expert trajectories $\tau_e \sim \pi_{\text{expert}}$ with $\tau_e = (s_1,a_1,s_2,a_2,\ldots)$
        \REQUIRE Learning rates $\alpha_w,\alpha_\theta$, entropy weight $\lambda$, PPO clip $\epsilon$
        \ENSURE Policy $\pi_\theta$ that imitates $\pi_{\text{expert}}$

        \STATE Initialize policy parameters $\theta$ and discriminator parameters $w$
        \FOR{\textbf{iter} $=1,2,\ldots$}
            \STATE \textbf{Collect on-policy data:} roll out $\pi_\theta$ to obtain trajectories $\tau_\theta$
            \STATE \textbf{Update discriminator} $d_w(s,a)$:
            \STATE \hspace{1em}$\displaystyle
                \delta_w \leftarrow 
                \sum_{(s,a)\in\tau_e}\nabla_w \log d_w(s,a)
                + \sum_{(s,a)\in\tau_\theta}\nabla_w \log\!\big(1-d_w(s,a)\big)$
            \STATE \hspace{1em}$w \leftarrow w + \alpha_w \,\delta_w$ \hfill $\vartriangleright$ \textit{Binary cross-entropy step}
            \STATE \textbf{Form imitation reward} from discriminator:
            \STATE \hspace{1em}$r_D(s,a) \leftarrow -\log\!\big(1-d_w(s,a)\big)$ \hfill $\vartriangleright$ \textit{GAIL reward}
            \STATE Estimate returns $\hat{G}_t$ and advantages $\hat{A}_t$ from $r_D$ (e.g., GAE)
            \STATE \textbf{Update policy} with PPO:
            \STATE \hspace{1em}For minibatches $\mathcal{M} \subset \tau_\theta$:
            \STATE \hspace{2em}$\displaystyle
                L^{\text{CLIP}}(\theta) =
                \mathbb{E}_{(s,a)\in\mathcal{M}}
                \Big[
                  \min\big(
                    \rho_\theta \hat{A},
                    \operatorname{clip}(\rho_\theta,1-\epsilon,1+\epsilon)\hat{A}
                  \big)
                \Big]
                - \lambda\, \mathbb{E}_{s\in\mathcal{M}} \big[ H(\pi_\theta(\cdot|s)) \big]$
            \STATE \hspace{2em}where $\rho_\theta = \frac{\pi_\theta(a|s)}{\pi_{\theta_{\text{old}}}(a|s)}$
            \STATE \hspace{1em}$\theta \leftarrow \theta + \alpha_\theta \,\nabla_\theta L^{\text{CLIP}}(\theta)$
        \ENDFOR
        \STATE \textbf{return} $\pi_\theta$
    \end{algorithmic}
\end{algorithm}

\end{document}